\documentclass[runningheads]{llncs}
\usepackage{graphicx}

\usepackage{tikz}
\usepackage{comment} 
\usepackage{amsmath,amssymb} 
\usepackage{color}

\usepackage{array}
\newcolumntype{C}[1]{>{\centering\let\newline\\\arraybackslash\hspace{0pt}}m{#1}}
\usepackage{duckuments}

\usepackage{url}
\usepackage{mwe}
\usepackage{multirow}
\usepackage{subcaption} 
\usepackage{epsfig}
\usepackage[linesnumbered,boxed,ruled,commentsnumbered]{algorithm2e} 

\newcommand{\TITLE}{
AIM 2020 Challenge on Image Extreme Inpainting
}
\newcommand{\fref}[1]{Fig.~\ref{#1}}
\newcommand{\tref}[1]{Table~\ref{#1}}

\newcommand{\sref}[1]{Section~\ref{#1}}
\newcommand{\rulesep}{\unskip\ \vrule\ }
\title{\TITLE}

\makeatletter
\DeclareRobustCommand\onedot{\futurelet\@let@token\@onedot}
\def\@onedot{\ifx\@let@token.\else.\null\fi\xspace}
\def\eg{\emph{e.g}\onedot} 
\def\ie{\emph{i.e}\onedot}

\def\etal{\emph{et al}\onedot}
\makeatother

\usepackage{tabularx}

\usepackage{array}
\newcolumntype{P}[1]{>{\centering\arraybackslash}p{#1}}

\begin{document}
\pagestyle{headings}
\mainmatter
\def\ECCVSubNumber{100}  

\titlerunning{AIM 2020 Challenge on Image Extreme Inpainting}
%
	
\author{Evangelos Ntavelis \and Andr\'es Romero \and Siavash Bigdeli \and Radu Timofte \and 
Zheng Hui \and Xiumei Wang \and Xinbo Gao \and 
Chajin Shin \and Taeoh Kim \and Hanbin Son \and Sangyoun Lee \and 
Chao Li \and Fu Li \and Dongliang He \and Shilei Wen \and Errui Ding \and 
Mengmeng Bai \and Shuchen Li \and 
Yu Zeng \and Zhe Lin \and Jimei Yang \and Jianming Zhang \and Eli Shechtman \and Huchuan Lu \and 
Weijian Zeng \and Haopeng Ni \and Yiyang Cai \and Chenghua Li \and 
Dejia Xu \and Haoning Wu \and Yu Han \and 
Uddin S. M. Nadim \and Hae Woong Jang \and Soikat Hasan Ahmed \and Jungmin Yoon \and Yong Ju Jung \and 
Chu-Tak Li \and Zhi-Song Liu \and Li-Wen Wang \and Wan-Chi Siu \and Daniel P.K. Lun \and 
Maitreya Suin \and Kuldeep Purohit \and A. N. Rajagopalan \and 
Pratik Narang \and Murari Mandal \and Pranjal Singh Chauhan 
}
\authorrunning{E. Ntavelis, A. Romero, S. Bigdeli, R. Timofte \etal}
%
\institute{}
\maketitle

\begin{abstract}
This paper reviews the AIM 2020 challenge on extreme image inpainting. This report focuses on proposed solutions and results for two different tracks on extreme image inpainting: classical image inpainting and semantically guided image inpainting. The goal of track 1 is to inpaint large part of the image with no supervision. Similarly, the goal of track 2 is to inpaint the image by having access to the entire semantic segmentation map of the input. The challenge had 88 and 74 participants, respectively. 11 and 6 teams competed in the final phase of the challenge, respectively. This report gauges current solutions and set a benchmark for future extreme image inpainting methods.
\end{abstract}
\let\thefootnote\relax\footnotetext{
E. Ntavelis (entavelis@ethz.ch, ETH Zurich and CSEM SA), A. Romero, S. Bigdeli, and R. Timofte are the AIM 2020 challenge organizers, while the other authors participated in the challenge.\\
Appendix A contains the authors’ teams and affiliations.\\
AIM webpage: \url{http://www.vision.ee.ethz.ch/aim20/} \\
Github webpage: \url{https://github.com/vglsd/AIM2020-Image-Inpainting-Challenge}
}
\begin{figure}[t]
	\centering%
	\newcommand{\wid}{.24\linewidth}
	
	\begin{tabular}{|@{}*{1}{P{\textwidth}@{}}|}
	\hline
    Track 1 
    \vspace{-0.1cm}
	\end{tabular}
	\begin{tabular}{|ccc||c|}
         
	     \includegraphics[width=\wid]{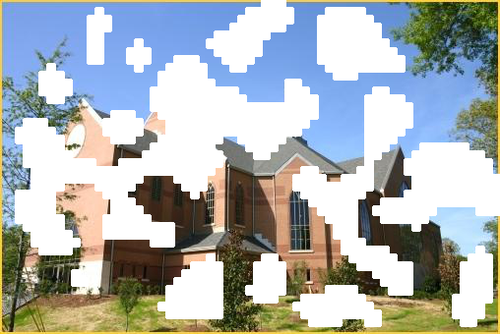} &
	     \includegraphics[width=\wid]{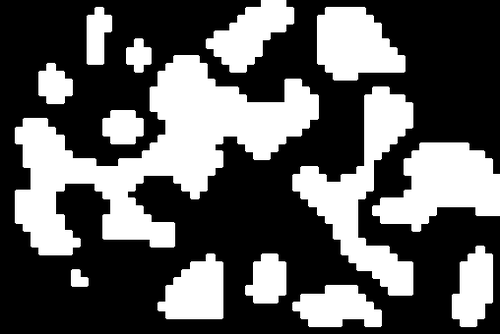} &
	     \hspace{\wid} &
	     \includegraphics[width=\wid]{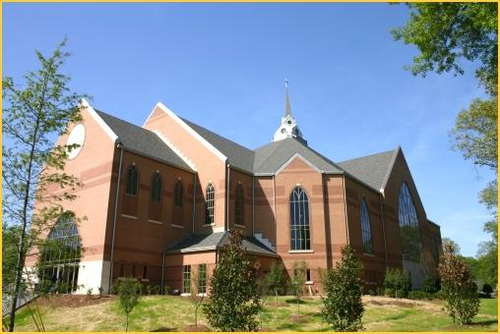} \\
	     
	     \includegraphics[width=\wid]{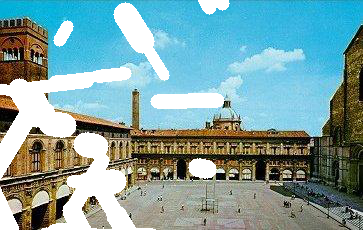} &
	     \includegraphics[width=\wid]{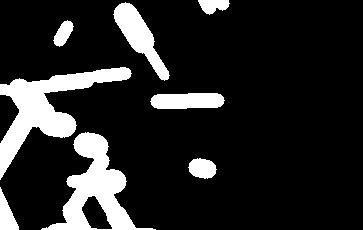} &
	     \hspace{\wid} &
	     \includegraphics[width=\wid]{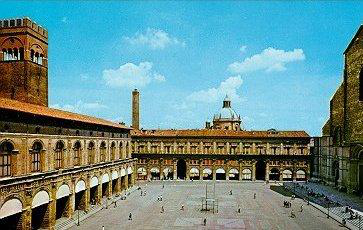} \\	     

    \end{tabular}
    \rulesep
    \rulesep
	\begin{tabular}{|@{}*{1}{P{\textwidth}@{}}|}
	\hline
    Track 2
    \vspace{-0.1cm}
	\end{tabular}
	\begin{tabular}{|ccc||c|}

	     \includegraphics[width=\wid]{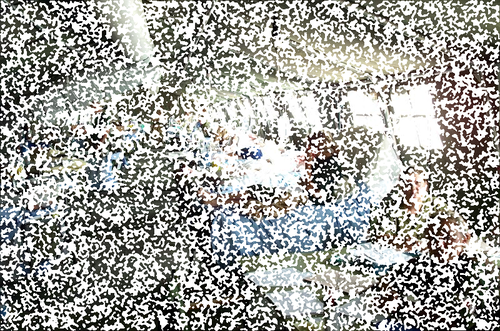} &
	     \includegraphics[width=\wid]{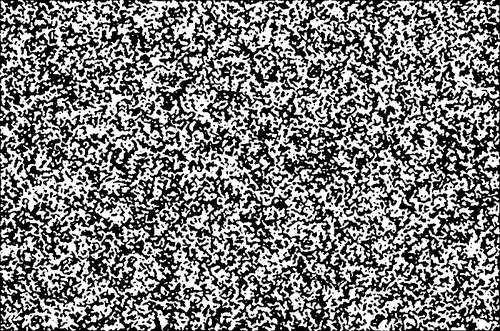} &
	     \includegraphics[width=\wid]{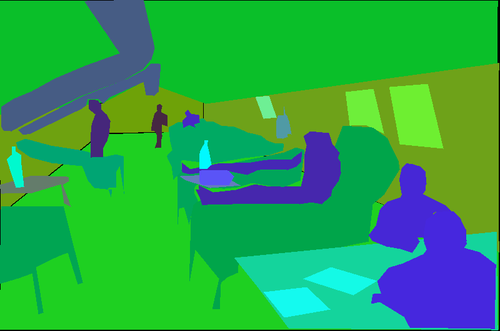} &
	     \includegraphics[width=\wid]{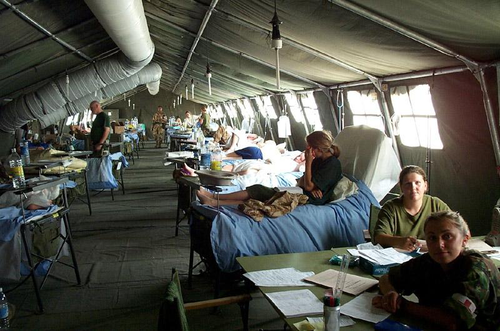} \\
	     
	     \includegraphics[width=\wid]{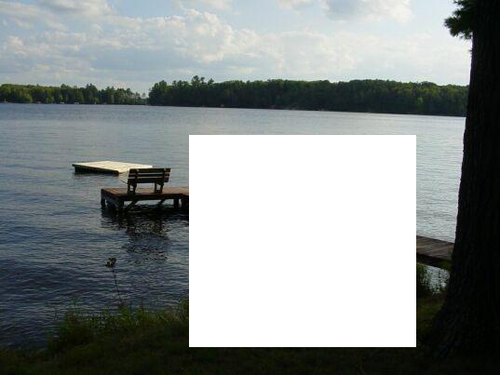} &
	     \includegraphics[width=\wid]{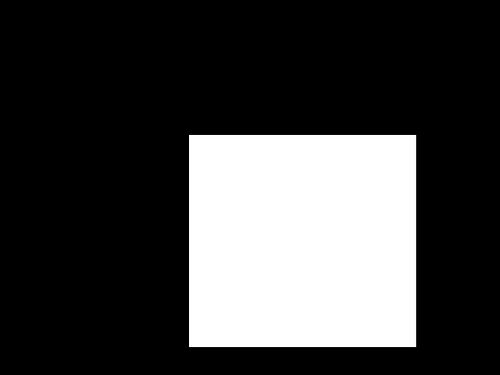} &
	     \includegraphics[width=\wid]{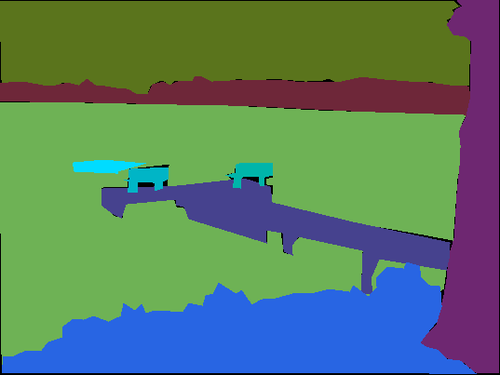} &
	     \includegraphics[width=\wid]{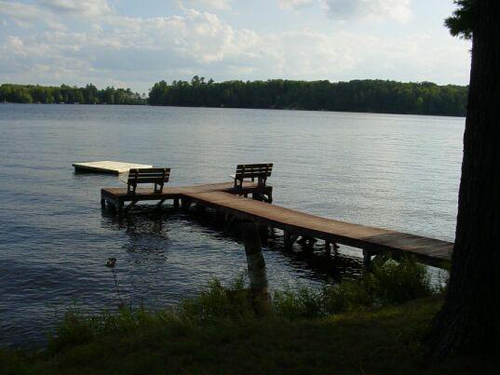} \\	     
	\hline
	\end{tabular}
	\begin{tabular}{@{}*{4}{P{.25\textwidth}@{}}}
    Input RGB & Input Mask & Input Seg & Ground-Truth	
	\end{tabular}
	\caption{Visual example of the extreme degraded input images and ground truth images used in the challenge. In both tracks, we automatically degraded each image to simulate different levels of extreme degradation. In Track 1 the aim is to produce realistic images using the Input RGB and Input Mask, and no additional information. Track 2 uses in addition a semantic segmentation map (Input Seg) as condition information for the generation of well defined objects.}
	\label{fig:intro}
\end{figure}

\section{Introduction}
Image inpainting is the task of recovering regions with some level of corrupted or missing regions. Normally, these regions are the outcome of degradation, artifacts, or they were altered by human intervention such as whitening object removal or image manipulation. The goal of this task is to fill in the missing pixels of the image, so the generated pixels are harmonious and perceptually plausible looking with the rest of the image.

Since the introduction of Generative Adversarial Networks (GANs)~\cite{goodfellow2014generative}, recent efforts on image inpaiting have produced impressive results for replacing objects~\cite{yu2018generative,yu2019free}, retouching landscapes~\cite{SPADE,Bau_Ganpaint_2019}, or altering the content of a scene~\cite{hong2018learning,ntavelis2020sesame}, either by given weak supervision (semantic segmentation guiding in addition to the binary mask to reconstruct) or no supervision (only using binary reconstruction mask). Interestingly, it is common to assume that the inpainted region is small (\eg squared bounding boxes in general) with respect to the entire image, which leads to expected high perceptual scores. 

In our \TITLE, we set the first extreme image inpainting challenge that aims at generating photo-realistic and perceptually appealing inpainted regions. The contestants are called to design a solution that is able to complete images of various sizes, spanning from low to high resolution, where the number of missing pixels is significant with respect to the entire image. Moreover, the masked regions can have a variety of shapes and sizes calling for an intricate approach to the problem. The objective of this challenge is to stimulate and propose a benchmark for further research in this direction.

This challenge is one of the AIM 2020 associated challenges on:
scene relighting and illumination estimation~\cite{elhelou2020aim_relighting}, image extreme inpainting~\cite{ntavelis2020aim_inpainting}, learned image signal processing pipeline~\cite{ignatov2020aim_ISP}, rendering realistic bokeh~\cite{ignatov2020aim_bokeh}, real image super-resolution~\cite{wei2020aim_realSR}, efficient super-resolution~\cite{zhang2020aim_efficientSR}, video temporal super-resolution~\cite{son2020aim_VTSR} and video extreme super-resolution~\cite{fuoli2020aim_VXSR}.

\begin{figure}[t]
    \centering
    \includegraphics[width=0.8\linewidth]{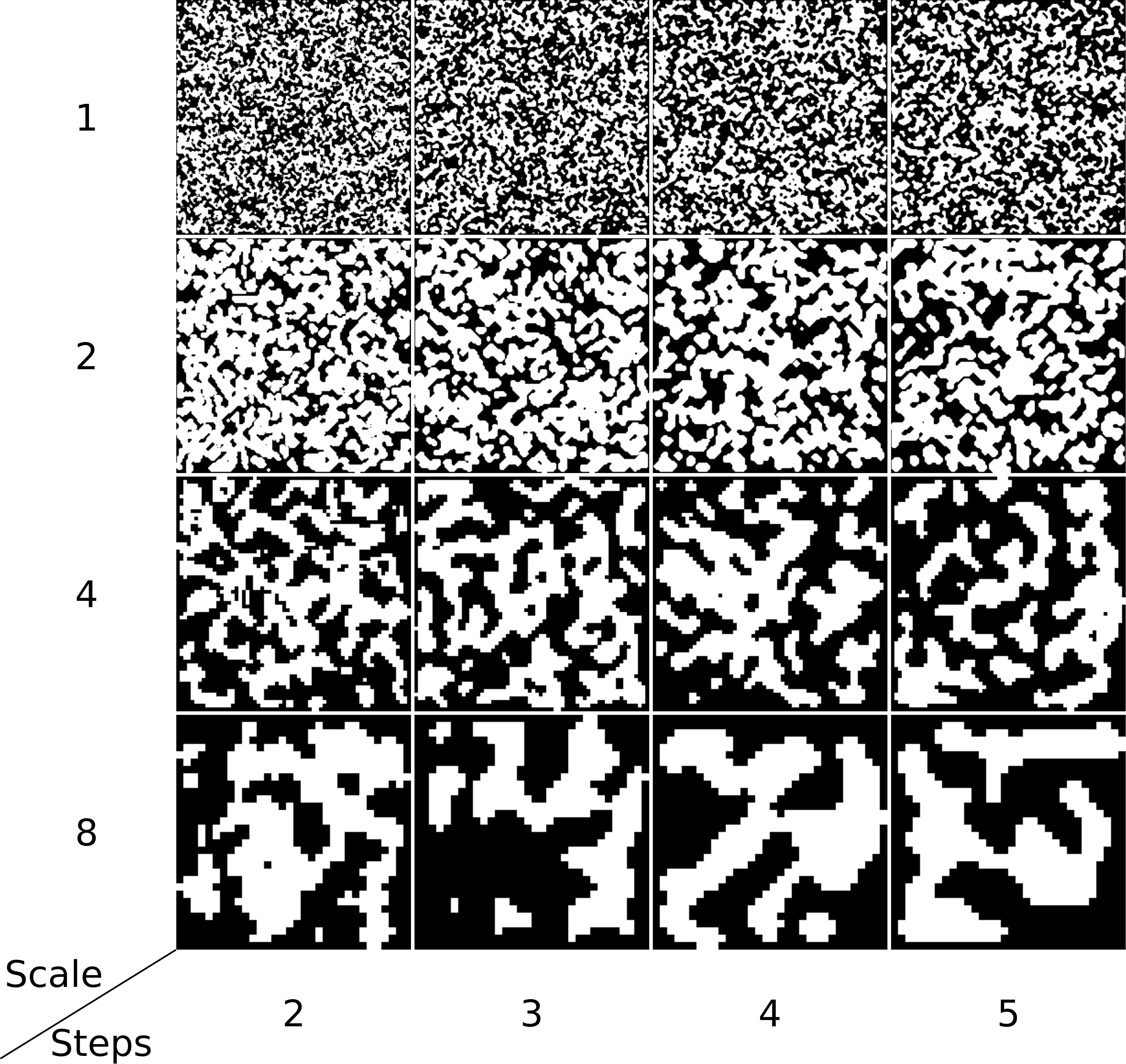}
    \caption{We use \textit{Cellural Automata} to generate the mask of an image. We apply a median filter at each step of the automaton to change its states. We start with a down-scaled mask size and re-scale after the automaton has reached its final state to create \textit{islands} of different sizes.}
    \label{fig:masks}
\end{figure}

\begin{table}[t]
    \centering
    \caption{Technical details of the participants. All teams that participated in Challenge 2 also participated in Challenge 1, and the runtime of using the semantic information is under mild assumptions the same}
    \label{tab:tech_details}    
    \newcommand{\sep}{~~}
            
    \begin{tabular}{c||c|c|c}
    \hline
    Team & Runtime (s/img) & GPU & Framework \\
    \hline
    \hline
    Rainbow & 0.2 & TITAN RTX & Pytorch \\
    Yonsei-MVPLab & 1.58 & RTX 2080Ti & Pytorch \\
    BossGao & 10 & Tesla V100 & Tensorflow \\
    ArtIst & 0.32 & GTX 1080Ti & Tensorflow \\
    DLUT & 36.39$^{*}$ & Tesla V100 & Pytorch \\
    AI-Inpainting & 3.54 & RTX 2080Ti & Pytorch \\
    qwq & 2.5 & GTX 1080Ti & Pytorch \\
    CVIP Inpainting & 0.57 & RTX 2080Ti & Pytorch \\
    DeepInpaintingT1 & 6.4 & RTX 2080Ti & Pytorch \\
    IPCV\_IITM & 1.2 & Titan X & Tensorflow \\
    MultiCog & 0.44 & K40 & Keras \\
    \hline
    \end{tabular}
\end{table}

\footnotetext{\textsuperscript{*}Participant reported runtime in CPU time.}

\begin{table}[t]
    \centering
    \caption{Challenge results for Track 1: Source domain on the final test set}
    \label{tab:tr1-test-results}    
    \newcommand{\sep}{~~}
    \resizebox{\columnwidth}{!}
    {
            
    \begin{tabular}{c||c|c|c|c|c}
    \hline
    Team & FID$\downarrow$ & LPIPS$\downarrow$ & PSNR$\uparrow$ & SSIM$\uparrow$ & MAE$\uparrow$ \\
    \hline
    \hline
    Rainbow & 30.69 & 0.10 $\pm$ 0.07 & 26.71 $\pm$ 7.43 & 0.88 $\pm$ 0.09 & 0.03 $\pm$ 0.02 \\
    Yonsei-MVPLab & 30.71 & 0.11 $\pm$ 0.09 & 27.25 $\pm$ 8.09 & 0.89 $\pm$ 0.09 & 0.02 $\pm$ 0.02 \\
    BossGao & 31.23 & 0.11 $\pm$ 0.08 & 26.59 $\pm$ 8.37 & 0.88 $\pm$ 0.1 & 0.03 $\pm$ 0.02 \\
    ArtIst & 33.29 & 0.12 $\pm$ 0.08 & 26.64 $\pm$ 8.55 & 0.87 $\pm$ 0.1 & 0.03 $\pm$ 0.02 \\
    DLUT & 40.46 & 0.13 $\pm$ 0.09 & 26.15 $\pm$ 8.47 & 0.87 $\pm$ 0.1 & 0.03 $\pm$ 0.02 \\
    AiriaBeijingTeam & 40.63 & 0.13 $\pm$ 0.08 & 26.1 $\pm$ 6.91 & 0.87 $\pm$ 0.09 & 0.03 $\pm$ 0.02 \\
    qwq & 41.03 & 0.19 $\pm$ 0.13 & 25.95 $\pm$ 5.86 & 0.86 $\pm$ 0.11 & 0.03 $\pm$ 0.02 \\
    CVIP Inpainting Team & 44.29 & 0.16 $\pm$ 0.11 & 26.2 $\pm$ 7.59 & 0.87 $\pm$ 0.1 & 0.03 $\pm$ 0.02 \\
    DeepInpaintingT1 & 48.40 & 0.18 $\pm$ 0.11 & 26.64 $\pm$ 7.6 & 0.87 $\pm$ 0.09 & 0.03 $\pm$ 0.02 \\
    IPCV\_IITM & 93.95 & 0.30 $\pm$ 0.27 & 20.98 $\pm$ 7.61 & 0.68 $\pm$ 0.29 & 0.08 $\pm$ 0.08 \\
    MultiCog & 117.52 & 0.53 $\pm$ 0.18 & 17.58 $\pm$ 2.53 & 0.62 $\pm$ 0.14 & 0.09 $\pm$ 0.04 \\
    \hline
    \end{tabular}
    }
\end{table}

\begin{table}[t]
    \centering
    \caption{Challenge results for Track 2: Target domain on the final test set}
    \label{tab:tr2-test-results}    
    \newcommand{\sep}{~~}
            
    \begin{tabular}{c||c|c|c|c|c}
    \hline
    Team & FID$\downarrow$ & LPIPS$\downarrow$ & PSNR$\uparrow$ & SSIM$\uparrow$ & MAE$\uparrow$ \\
    \hline
    \hline    
    Rainbow & 32.60 & 0.11 $\pm$ 0.08 & 26.77 $\pm$ 7.82 & 0.88 $\pm$ 0.1 & 0.03 $\pm$ 0.02 \\
    ArtIst & 36.00 & 0.13 $\pm$ 0.08 & 27.11 $\pm$ 7.93 & 0.88 $\pm$ 0.09 & 0.03 $\pm$ 0.02 \\
    DLUT & 43.22 & 0.14 $\pm$ 0.10 & 25.8 $\pm$ 8.1 & 0.86 $\pm$ 0.1 & 0.03 $\pm$ 0.03 \\
    qwq & 43.38 & 0.19 $\pm$ 0.12 & 24.74 $\pm$ 5.43 & 0.85 $\pm$ 0.1 & 0.03 $\pm$ 0.03 \\
    DeepInpaintingT1 & 44.57 & 0.18 $\pm$ 0.11 & 26.94 $\pm$ 7.09 & 0.88 $\pm$ 0.09 & 0.03 $\pm$ 0.02 \\
    AI-Inpainting & 45.63 & 0.15 $\pm$ 0.10 & 25.79 $\pm$ 6.68 & 0.86 $\pm$ 0.1 & 0.03 $\pm$ 0.02 \\
    \hline
    \end{tabular}
\end{table}

\begin{figure}[t]
    \centering%
    \begin{tabular}{|c||ccc||c|}
    \hline
         \includegraphics[width=0.19\linewidth]{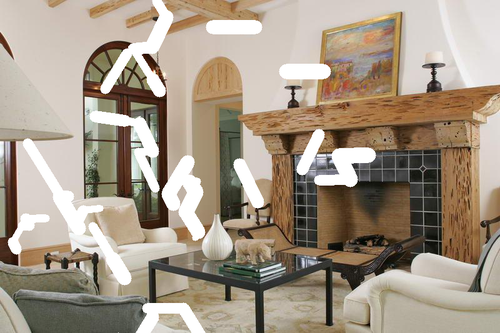} &
         \includegraphics[width=0.19\linewidth]{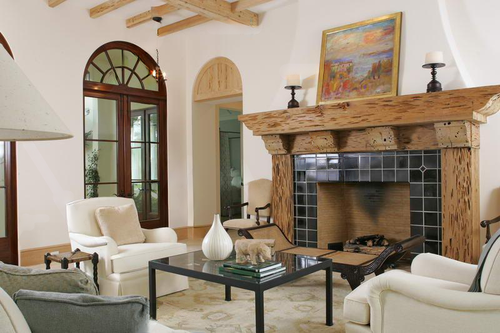} &
         \includegraphics[width=0.19\linewidth]{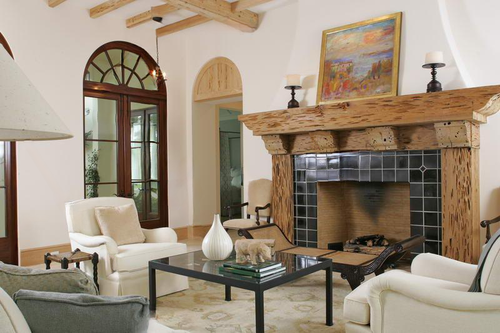} &
         \includegraphics[width=0.19\linewidth]{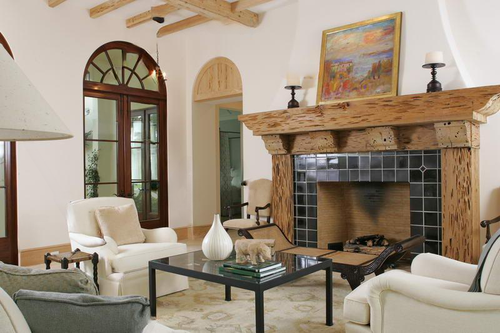} &
         \includegraphics[width=0.19\linewidth]{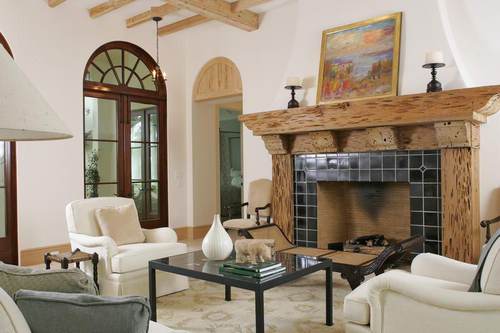} \\

         \includegraphics[width=0.19\linewidth]{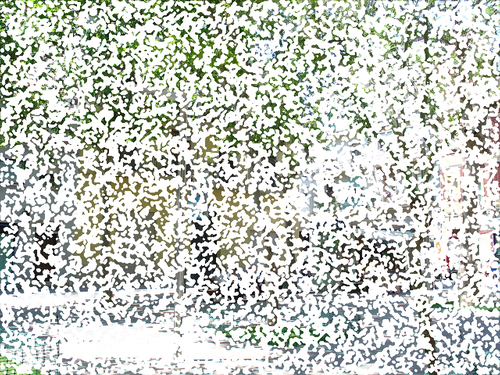} &
         \includegraphics[width=0.19\linewidth]{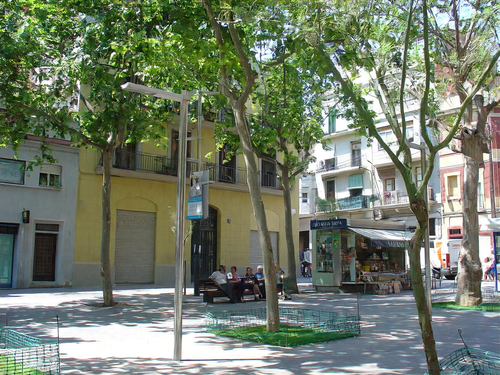} &
         \includegraphics[width=0.19\linewidth]{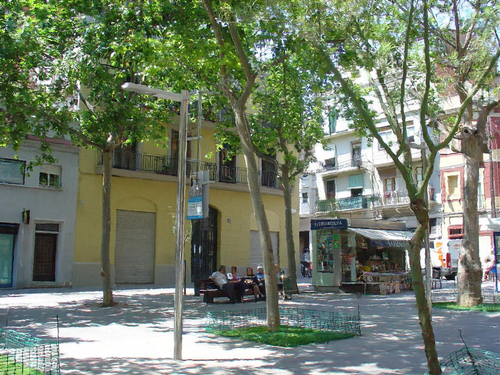} &
         \includegraphics[width=0.19\linewidth]{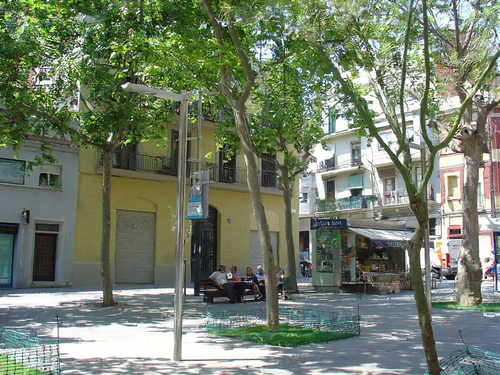} &
         \includegraphics[width=0.19\linewidth]{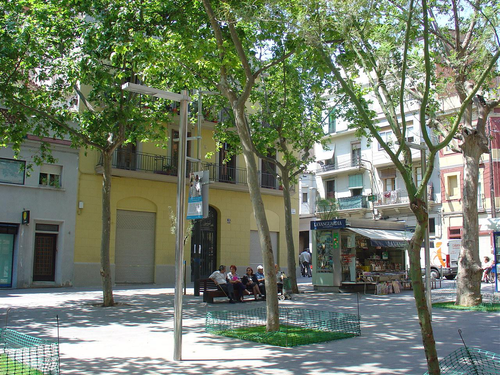} \\

         \includegraphics[width=0.19\linewidth]{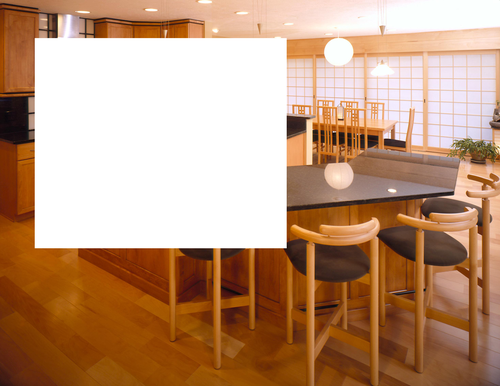} &
         \includegraphics[width=0.19\linewidth]{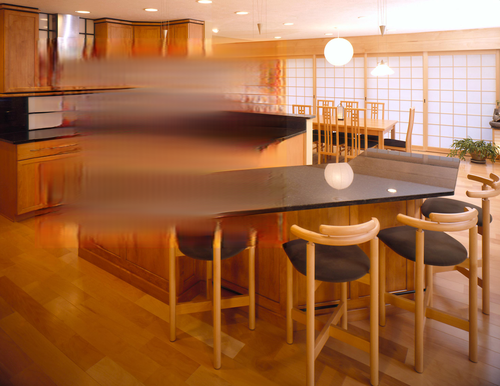} &
         \includegraphics[width=0.19\linewidth]{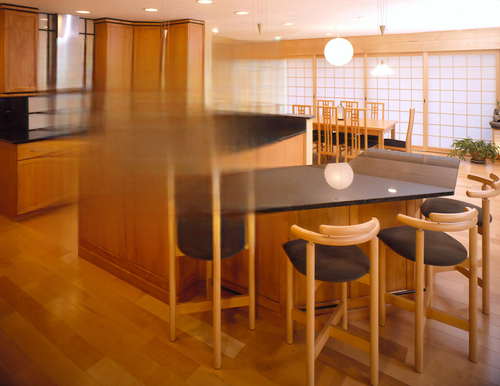} &
         \includegraphics[width=0.19\linewidth]{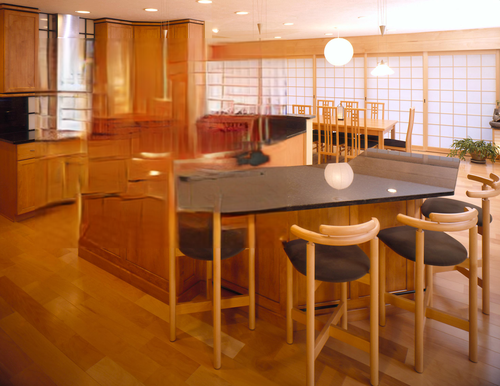} &
         \includegraphics[width=0.19\linewidth]{Figures/Comparison_T1/AIM_IC_t1_test_47.png} \\
         
         \includegraphics[width=0.19\linewidth]{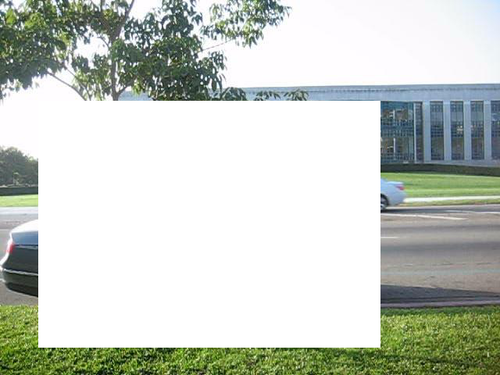} &
         \includegraphics[width=0.19\linewidth]{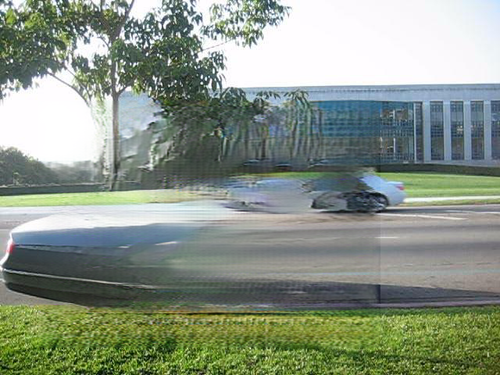} &
         \includegraphics[width=0.19\linewidth]{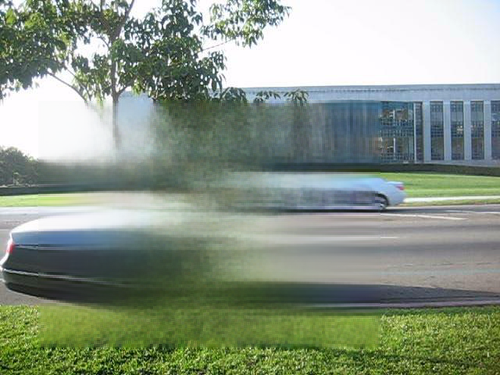} &
         \includegraphics[width=0.19\linewidth]{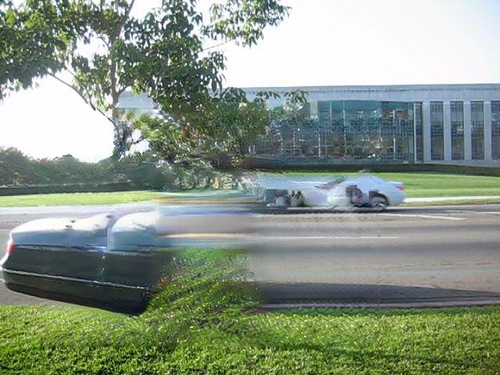} &
         \includegraphics[width=0.19\linewidth]{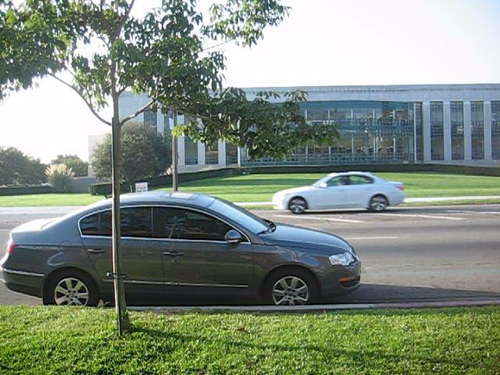} \\         
 
    \hline
    \end{tabular}
    \begin{tabular}{@{}*{5}{P{.2\textwidth}@{}}}
    Input & Rainbow & Yonsei-MVPlab & BossGao & Ground-Truth    
    \end{tabular}
    \caption{Qualitative comparison over the top three methods in Track 1. Rainbow and Yonsei-MVPlab team models can successfully interpret the local scene textures and propagate information to the missing regions when the inpainted region is not too extreme. Overall, the results from the rainbow team has fewer visual artifacts. Zoom in for better visual comparisons.}
    \label{fig:qualitative_track1}
\end{figure}

\begin{figure}[t]
    \centering%
    \begin{tabular}{|cc||ccc||c|}
    \hline
         \includegraphics[width=0.16\linewidth]{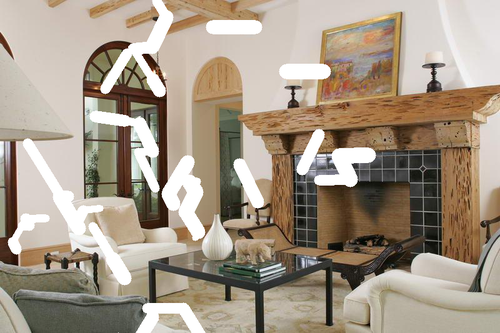} &
         \includegraphics[width=0.16\linewidth]{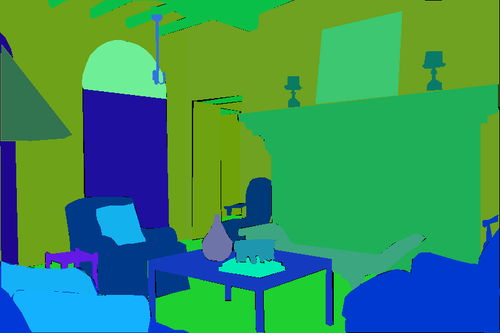} &
         \includegraphics[width=0.16\linewidth]{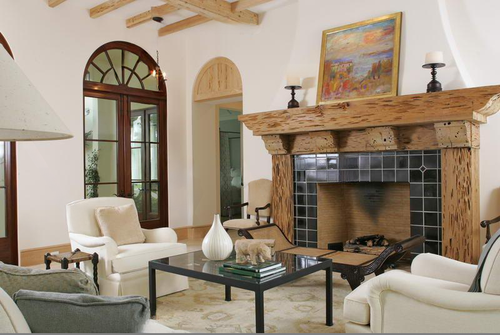} &
         \includegraphics[width=0.16\linewidth]{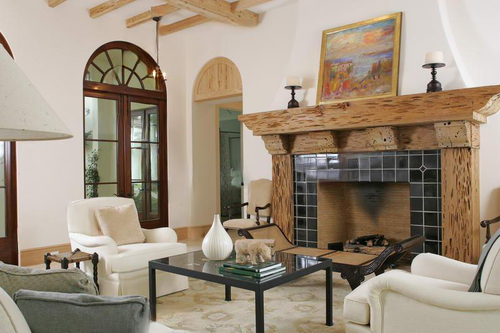} &
         \includegraphics[width=0.16\linewidth]{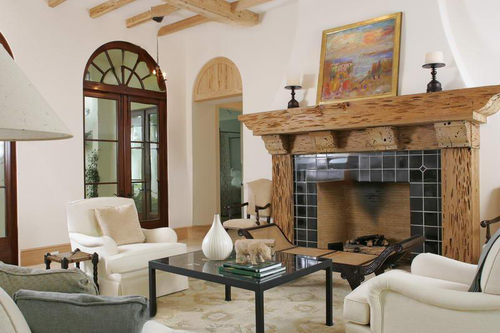} &
         \includegraphics[width=0.16\linewidth]{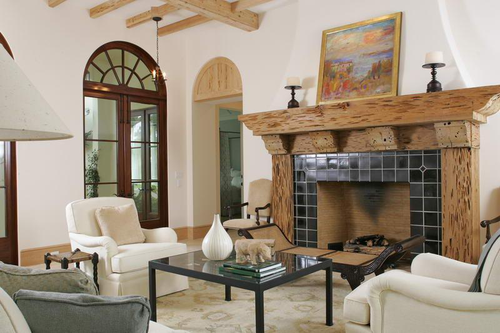} \\               

         \includegraphics[width=0.16\linewidth]{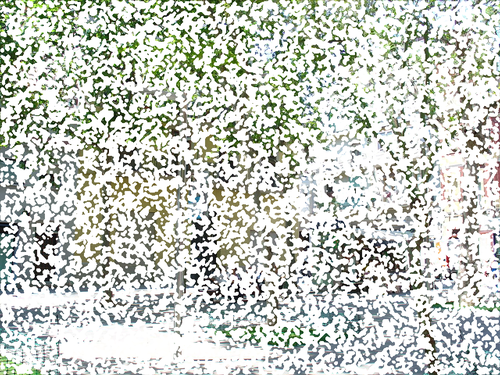} &
         \includegraphics[width=0.16\linewidth]{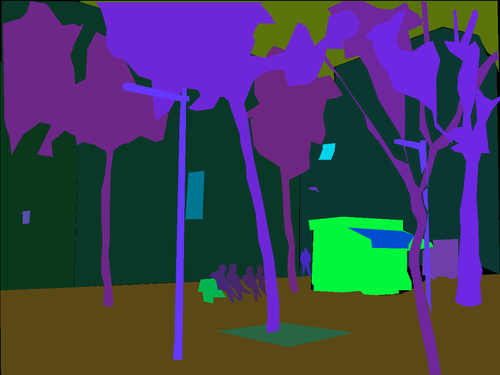} &
         \includegraphics[width=0.16\linewidth]{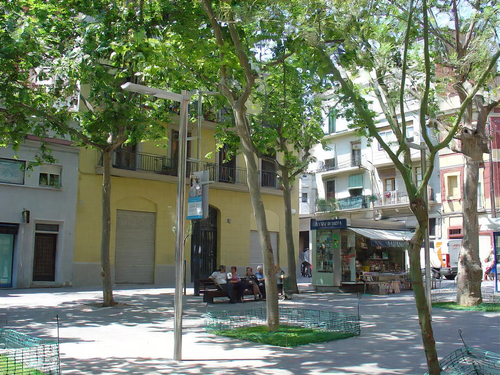} &
         \includegraphics[width=0.16\linewidth]{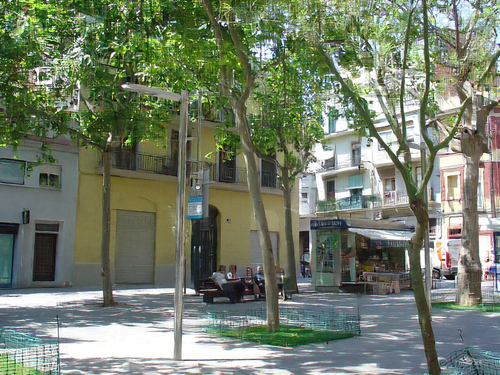} &
         \includegraphics[width=0.16\linewidth]{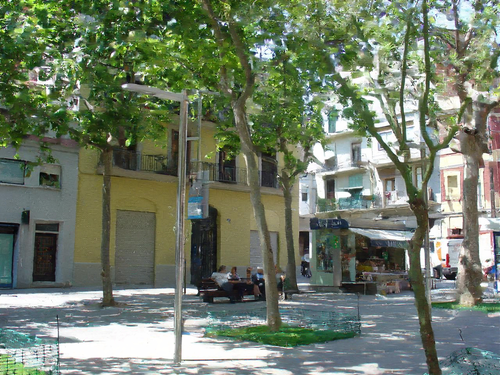} &
         \includegraphics[width=0.16\linewidth]{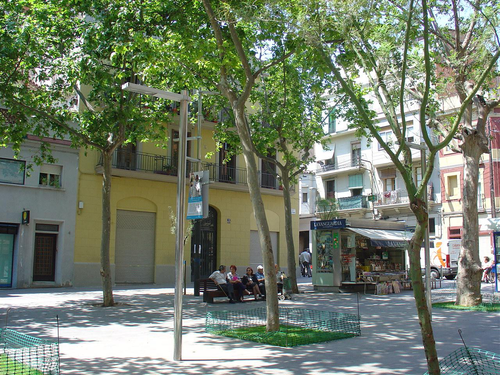} \\               
         
         \includegraphics[width=0.16\linewidth]{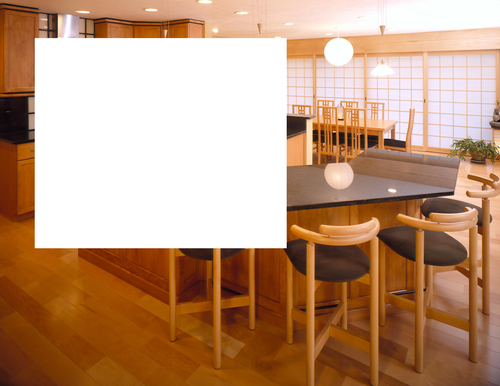} &
         \includegraphics[width=0.16\linewidth]{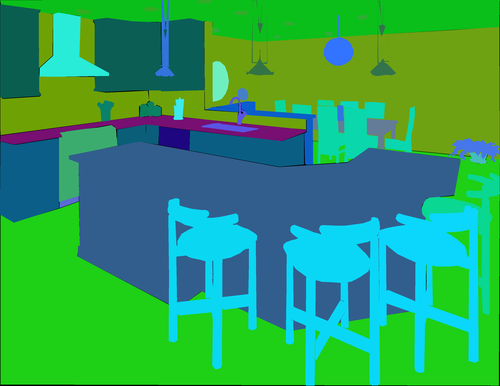} &
         \includegraphics[width=0.16\linewidth]{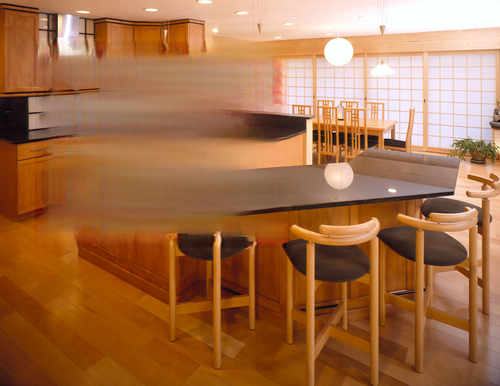} &
         \includegraphics[width=0.16\linewidth]{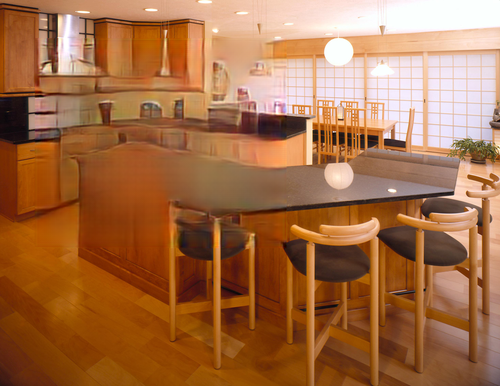} &
         \includegraphics[width=0.16\linewidth]{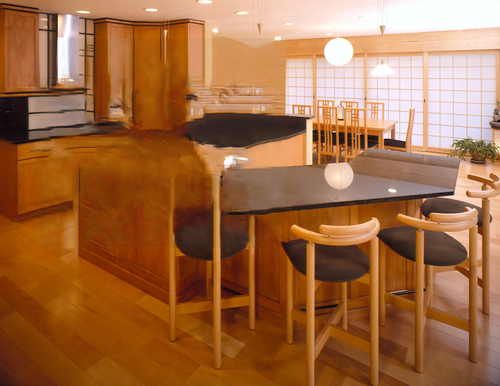} &
         \includegraphics[width=0.16\linewidth]{Figures/Comparison_t2/AIM_IC_t2_test_47.png} \\               
         
         \includegraphics[width=0.16\linewidth]{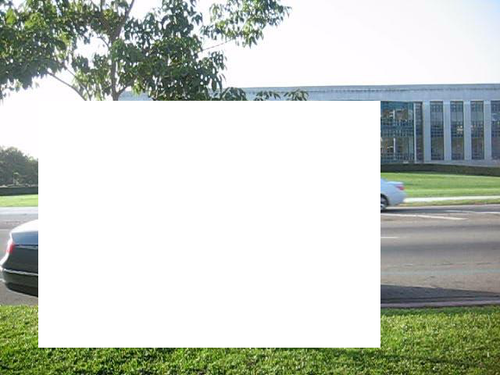} &
         \includegraphics[width=0.16\linewidth]{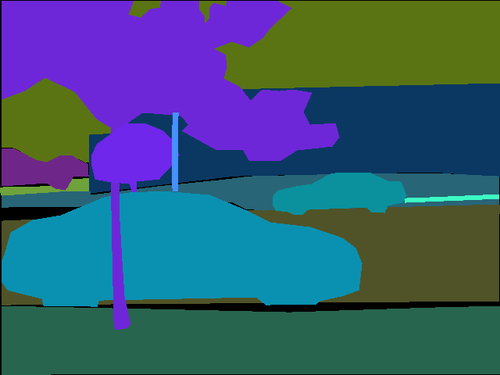} &
         \includegraphics[width=0.16\linewidth]{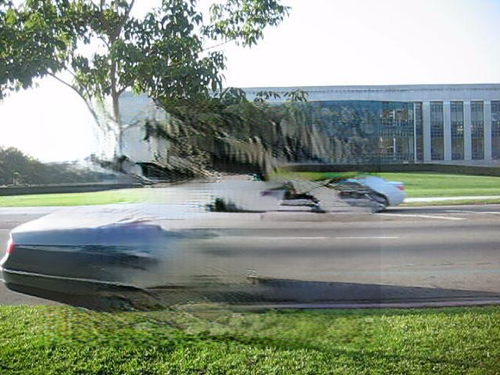} &
         \includegraphics[width=0.16\linewidth]{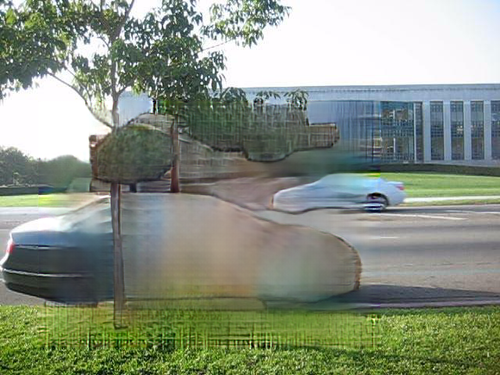} &
         \includegraphics[width=0.16\linewidth]{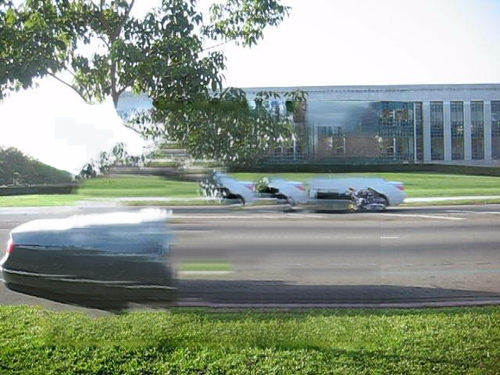} &
         \includegraphics[width=0.16\linewidth]{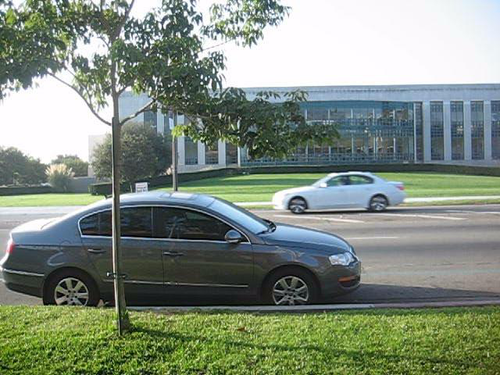} \\               
         
         \includegraphics[width=0.16\linewidth]{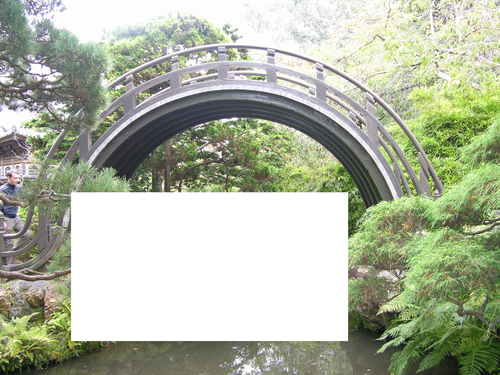} &
         \includegraphics[width=0.16\linewidth]{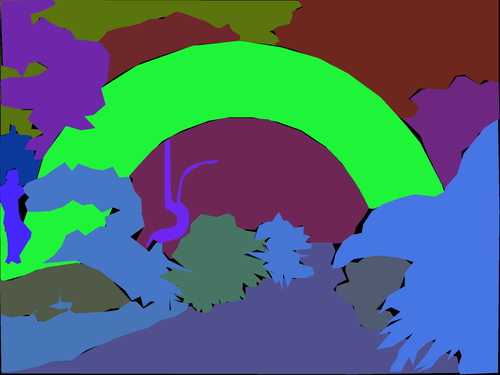} &
         \includegraphics[width=0.16\linewidth]{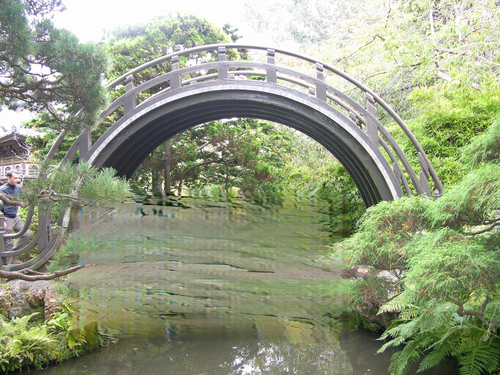} &
         \includegraphics[width=0.16\linewidth]{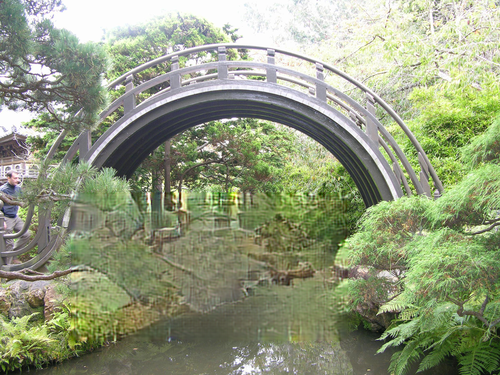} &
         \includegraphics[width=0.16\linewidth]{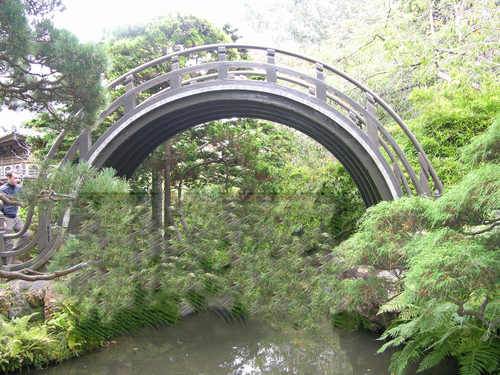} &
         \includegraphics[width=0.16\linewidth]{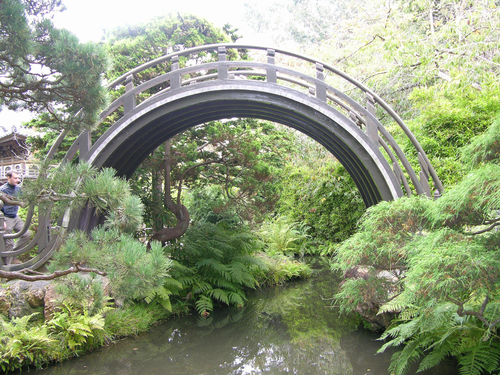} \\
 
    \hline
    \end{tabular}
    \begin{tabular}{@{}*{6}{P{.1666\textwidth}@{}}}
    Input RGB & Input Seg & Rainbow & ArtIst & DLUT & Ground-Truth    
    \end{tabular}
    \caption{Qualitative comparison over the top three methods in Track 2. Semantic labels help the model from ArtIst to better locate object boundaries. Although rainbow team yields better perceptual performance, it is unable to assign consistent object boundaries. We depict the same images as in \fref{fig:qualitative_track1} for comparison in both tracks.}
    \label{fig:qualitative_track2}
    
\end{figure}

\begin{table}[t]
    \centering
    \caption{Results per mask type for three top solutions of Track 1}
    \label{tab:t1maskcomp}    
    {
    \begin{tabular}{c|c||c|c|c|c}
    \hline
    Team & Mask Type & LPIPS$\downarrow$ & PSNR$\uparrow$ & SSIM$\uparrow$ & MAE$\uparrow$ \\
    
    \hline
 & Box & 0.15 & 22.49 & 0.82 & 0.04 \\
Rainbow & Cellural Automata & 0.12 & 25.72 & 0.89 & 0.03 \\
 & Free-Form & 0.05 & 32.44 & 0.95 & 0.01 \\
    \hline
 & Box & 0.17 & 23.45 & 0.83 & 0.03 \\
Yonsei-MVPLab & Cellural Automata & 0.13 & 25.40 & 0.88 & 0.03 \\
 & Free-Form & 0.04 & 33.44 & 0.95 & 0.01 \\
    \hline
 & Box & 0.15 & 22.09 & 0.81 & 0.04 \\
BossGao & Cellural Automata & 0.13 & 25.31 & 0.87 & 0.03 \\
 & Free-Form & 0.05 & 32.94 & 0.95 & 0.01 \\
    \hline
 & Box & 0.30 & 11.47 & 0.66 & 0.14 \\
Masked Images & Cellural Automata & 0.66 & 8.21 & 0.23 & 0.23 \\
 & Free-Form & 0.25 & 16.02 & 0.70 & 0.07 \\
    \hline
    
    \end{tabular}
    }
\end{table}

\begin{table}[t]
    \centering
    \caption{Results per mask type for three top solutions of Track 2}
    \label{tab:t2maskcomp}    
    {
    \begin{tabular}{c|c||c|c|c|c}
    \hline
    Team & Mask Type & LPIPS$\downarrow$ & PSNR$\uparrow$ & SSIM$\uparrow$ & MAE$\uparrow$ \\
    \hline
 & Box & 0.16 & 22.15 & 0.81 & 0.04\\
Rainbow & Cellural Automata & 0.10 & 26.85 & 0.90 & 0.02\\
 & Free-Form & 0.05 & 32.84 & 0.95 & 0.01\\
    \hline
 & Box & 0.16 & 23.07 & 0.83 & 0.03\\
ArtIst & Cellural Automata & 0.16 & 25.99 & 0.88 & 0.03\\
 & Free-Form & 0.06 & 33.08 & 0.95 & 0.01\\
    \hline
 & Box & 0.16 & 21.70 & 0.81 & 0.04\\
DLUT & Cellural Automata & 0.18 & 24.49 & 0.85 & 0.03\\
 & Free-Form & 0.07 & 32.06 & 0.94 & 0.02\\
    \hline
 & Box & 0.32 & 11.16 & 0.63 & 0.14\\
Masked Images & Cellural Automata & 0.65 & 8.77 & 0.26 & 0.22\\
 & Free-Form & 0.27 & 15.61 & 0.69 & 0.09 \\
    \hline
    \end{tabular}
    }
\end{table}

\begin{figure}[t]
    \centering%
     \includegraphics[width=1.\linewidth]{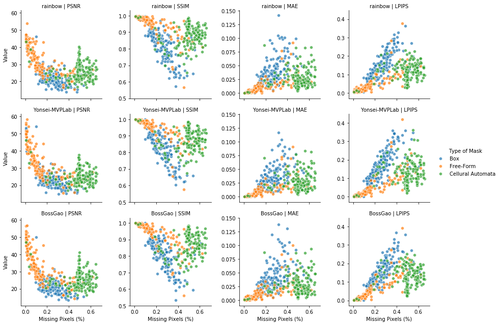}
    \caption{A visualization of the performance of the top three methods for Track 1 in relation to the percentage of the missing pixels in the input image. Each individual point corresponds to an image of the test set. For PSNR and SSIM larger values are better while for MAE and LPIPS smaller values indicate better performance. For all methods we observe that while \textit{cellural automata} usually remove more pixels than the \textit{box} masks, the latter are difficult to inpaint and consistenly produce worse results for all metrics. The easiest type of mask to inpaint is the Free-Form mask. The missing pixels are less in number and not concentrated as in the \textit{box} case}
    \label{fig:task1post}
\end{figure}

\begin{figure}[t]
    \centering%
     \includegraphics[width=1.\linewidth]{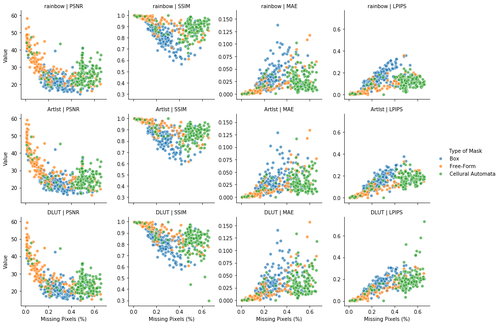}
    \caption{A visualization of the performance of the top three methods for Track 2 in relation to the percentage of the missing pixels in the input image. Each individual point corresponds to an image of the test set. For PSNR and SSIM larger values are better while for MAE and LPIPS smaller values indicate better performance. We make similar observations with Track 1, as described in \fref{fig:task1post}}
    \label{fig:task2post}
\end{figure}

\begin{figure}[t]
    \centering%
     \includegraphics[width=1.\linewidth]{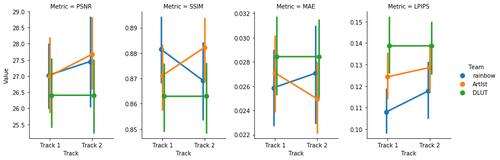}
    \caption{Point plot of the performance of the top three teams that participated in both tracks, calculated in a subset of the test set shared between them. \textit{DLUT} submitted the same model for both tracks, without utilizing the additional information. We observe that the top two teams produce worse perceptual results (LPIPS) when using semantics. For the second track, the first team is worse for the majority of metrics while the second team is better compared to the results of the first track, indicating that the semantic information can be both beneficial and detrimental to the inpainting task, based on how it is integrated into the network.
    }
    \label{fig:taskcomparison}
\end{figure}

\section{AIM 2020 Challenge}
The goals of the \TITLE are: \textit{(i)} to advance towards more challenging conditions for image inpainting methods, \textit{(ii)} to provide a common benchmark, protocol and dataset for image inpainting methods, and \textit{(iii)} to establish current state-of-the-art under extreme conditions. The aim is to obtain a network design / solution capable of producing high quality results with the best perceptual quality and similarity to the reference ground truth.

\subsection{Description}
In the classical sense, the task of image inpainting aims to reconstruct \textit{damaged} or \textit{missing} areas of an image. The ideal reconstruction provided by an inpainting operation should recreate the original pixels. However, in many cases all the information regarding a present entity in an image, \ie an object, is vanished behind the missing pixels. It is impossible for a model to perfectly reconstruct the missing information without external guidance. Motivated by this problem we created two tracks for the AIM 2020 Extreme Inpainting Challenge: (1) the classical image inpainting track, where no additional information is used, and (2) the semantically guide image inpainting track, where the pixel-level semantic labels of the whole image are provided to drive the generation of the missing pixels.  

\subsection{Dataset}
We use a partition of ADE20k dataset~\cite{zhou2017scene} for both tracks of the challenge. The ADE20k is a dataset with a large diversity of contents. A public training and validation set is provided, with each image being fully annotated. The resolution of the images provided in the dataset is highly diverse, and for our challenge, the inpainting must be achieved in the full scale of the image.

We aimed to provide images for both tasks that are drawn from the same distribution. Moreover, for the semantically guided challenge, we decided not to include all the classes and bypass the problem of the distribution's long tail. To achieve this, we selected the union of (1) the most occurring semantic classes per image and (2) the most occurring semantic classes per pixel. Ultimately, we ended up with 51 semantic classes and we filtered the images so that at least 90\% of the pixels in a particular image belong to these 51 semantic classes. The resulting training set was 10,330 to be used for both tracks.

In order to create the validation and test set of the challenge, we followed the same procedure as in the training set. We processed the validation set of ADE20k and divided the filtered images into validation and test subset. Each set was divided into three equal numbered parts: (1) images used solely for track 1, (2) images used solely for track 2, and (3) images used for both tracks.

We enriched our test set by mining pictures from the COCO-Stuff dataset~\cite{cocostuff}. We manually picked 200 images depicting scenes similar to the ones produced by our filtering process on ADE20k. Then, we defined a matching between the class labels of ADE20k and COCO-Stuff and translated the semantic maps accordingly. The additional COCO-Stuff images were divided into three groups similar to ADE20k images.

\subsection{Masking the images}
\label{sec:masks}
In order to push the boundaries of image inpainting, for this competition we are using three different types of masks: (\textit{i}) Rectangular masks with width and height between 30-70\% of each dimension, (\textit{ii}) brush strokes randomly drawn~\cite{yu2019free},
and (\textit{iii}) our own method generated masks based on cellural automata.

We propose this novel way of masking in order to create a distorted image that has holes distributed across its dimensions, while forming small \textit{islands} of missing pixels that are less trivial to fill compared to just salt and pepper noise. Cellural automata have been utilized in automatic content generation for games~\cite{Shaker2016}, but to our knowledge, this is the first time used for mask generation for image inpainting.

Our cellural automaton mask is a 2-dimensional grid of pixels having two states: either masked or not. We initialize our grid by randomly assigning a state to each pixel. At each step, every pixel's state is changed based on a majority vote on its neighborhood. We are using a Moor neighborhood: a 3-by-3 square, essentially applying a median filter. To facilitate creating islands of various sizes, we downscale the size of the mask before calculating the states of the automaton and re-scaling to the original size of the image to be masked. When re-scaling, we use the morphological operation of dilation to produce smoother edges. The resulted masks can be found in \fref{fig:masks}. To generate each mask, we randomly choose between down-scaling 1,2,4 or 8 times, and the number of applied steps (2-5).  

\subsection{Tracks}
\subsubsection{Track 1: Image Extreme Inpainting}
In this track, we provide only the degraded image with the corresponding mask to fill in the missing pixels. Formally, the aim of this track is to learn a mapping function $\mathbb{G}$ in order to produce a coherent and perceptually looking image $X = \mathbb{G}(X_{m}, M)$, where $X_{m}$ and $M$ are the degraded image and the mask image, respectively. See~\fref{fig:intro} upper row for a random example of inputs and ground-truth. For this track, only the images are to be used. The use of any other additional information (semantics, object instances) was explicitly not allowed.

\subsubsection{Track 2: Image Extreme Inpainting guided by pixel-wise semantic labels}
Here the task is also to complete the missing pixel information, but using a semantic image as guidance, so the generated image should resemble the same objects and shapes as in the semantic region. It can be formally depicted as $X = \mathbb{G}(X_{m}, M, S)$, where $S$ is the semantic map annotation corresponding to the ground-truth $X$. We depict an example of input images in~\fref{fig:intro} bottom row. For the semantic images, we provided a subset of the original ADE20k's semantic classes list, and the new number of classes is 51. The use of any other additional information, \eg object/instance information, was not allowed.

\subsection{Challenge Phases}
The challenge had three phases: (1) \textit{Development phase}: the
participants had access to training inputs and ground-truth, and the input images of the validation set. During this phase, the participants were free to select the number and type of masks described in \sref{sec:masks}.
(2) Validation phase: the participants had the opportunity to compute validation performance using PSNR and SSIM metrics by submitting their results on the server. A validation leaderboard was also available.
(3) Final test phase: the participants got access to the test images and had to submit their inpainted images along with the description of the method, code, and model weights for further reproducibility.

\section{Challenge Results}
For more than 150 participants registered on the two tracks combined, 11 teams entered the final phase of Track 1 and submitted results, code/executables, and factsheets. In Track 2, 6 teams entered the final phase. Track 2 participants also participated in Track 1. 
The Methods of the teams that entered the final phase are described in \sref{sec:methods}, the technical details of the teams in \tref{tab:tech_details}, and the team's affiliation is shown in \sref{sec:affiliation}.
\tref{tab:tr1-test-results}~and~\ref{tab:tr2-test-results} report the final results of Track 1 and 2 respectively, on the test data of the challenge. Noteworthy, for the top three methods in terms of both perceptual and fidelity metrics, we run a perceptual study to determine the winner of the challenge. Based on our results, Rainbow generated in overall more photo-realistic inpainted images than Yonsei-MVPLab and BossGao in Track 1 and than ArtIst and DLUT in Track 2, hence ranking 1st in both tracks of the first AIM Extreme Inpainting Challenge. We have to note that for Track 1 Yonsei-MVPLab performed better in the fidelity metrics.

\subsection{Architectures and main ideas}
All the proposed methods are GAN-based solutions. In most cases, a variant of PatchGAN~\cite{isola2016imagetoimage} is used for the discriminator, while many generators imitate the structure of recent state-of-the-art architectures with two Stage Architectures, one coarsely inpainting, and the other taking care of the finer details~\cite{yu2018generative,yu2019free}. The Edge-Connect approach \cite{nazeri2019edgeconnect} was also explored. Interestingly, the work of Hui~\etal~\cite{hui2020image}, the team winning this competition, was also employed by the second-best team of Track 1, Yonsei-MVPLab. The teams that performed the best in the competition propose novel components to their architectures.

\subsection{Handling of extreme image sizes}
Many of the participants reported that their solution could not handle high-resolution images without prompting an \textit{our of memory} error (OOM). Two approaches were mainly reported to tackle this issue. To circumvent this issue, many contestants downscaled the input images prior to inpainting and up-scaled the results with either a classical method, \eg bilinear upsampling, or using a Super Resolution network. This solution was only used by some teams to handle the big box holes. Alternatively, the input image was divided into patches, which after independent processing were stitched together to produce the final output.

\subsection{Handling of different mask types}

The three different mask types impose different problems and a few of the teams decided to handle them independently. We conducted an analysis on the results of the top three performing teams for each track,  which can be found in \tref{tab:t1maskcomp} and \ref{tab:t2maskcomp} respectively. As a comparison baseline, we show the metrics results when computed against the input image with holes. For the baseline case, the cellural automata masks remove the most pixels and thus produce the worst results, yet we can see that compared to the massive chunks of information removed from the box masks, cellural automata masks are easier to inpaint. In \fref{fig:task1post} and \ref{fig:task2post} we can observe how the top three teams of each track performed relative to the percentage of missing information.

\subsection{The effect of the semantic guidance}
In order to be able to fairly compare the results of the two tracks, a subset of their test set was shared. In \fref{fig:taskcomparison} we can see how the performance on a variety of metrics differs between the classical inpainting case, where no additional information is used, and the semantically guided case. The results indicate that semantic information can both increase and decrease the performance on the inpainting task, based on how its processing was implemented in the network. Some teams use the same network they used for the first track to produce results for the second one while other trivially incorporate the semantic information in their pipeline. These factors may contribute to counter-intuitive result that many teams produced worse results in the second track.

\subsection{Conclusions}
The \TITLE gauges the current solutions and proposes a benchmark for the problem of image extreme inpainting. Different methodologies and architectures have been proposed to tackle this problem. Most of these solutions built upon traditional image inpainting methods, which has been qualitative and quantitatively validated in this report that does not necessarily hold for the extreme case. Using semantic information as guidance does not always increase the performance of an image inpainting. We believe that \TITLE could boost research in this challenging direction.

As qualitative and quantitative results show, image extreme inpainting is an unexplored area, where state-of-the-art classical image inpainting methods fail to generalize. Remarkably, the number of missing pixels is not a critical factor for image inpainting, but the type of mask type.

\section{Challenge Methods and Teams}
\label{sec:methods}
\subsection{Rainbow}
\begin{figure}[t]
    \centering%
    \begin{minipage}{\textwidth}
        \begin{minipage}[c]{0.75\textwidth}
            \includegraphics[width=\textwidth]{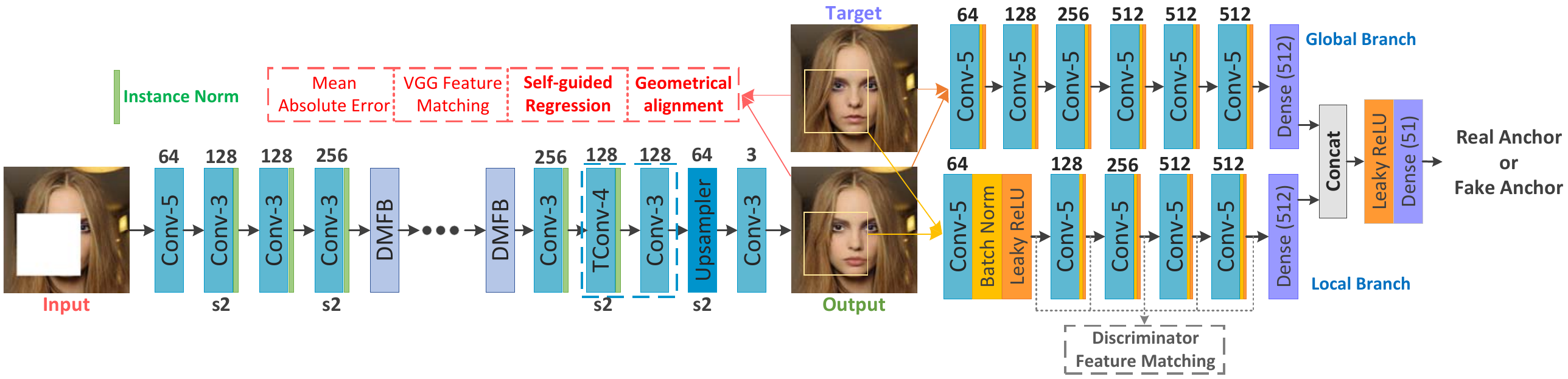} \subcaption{}
            
            \includegraphics[width=\textwidth]{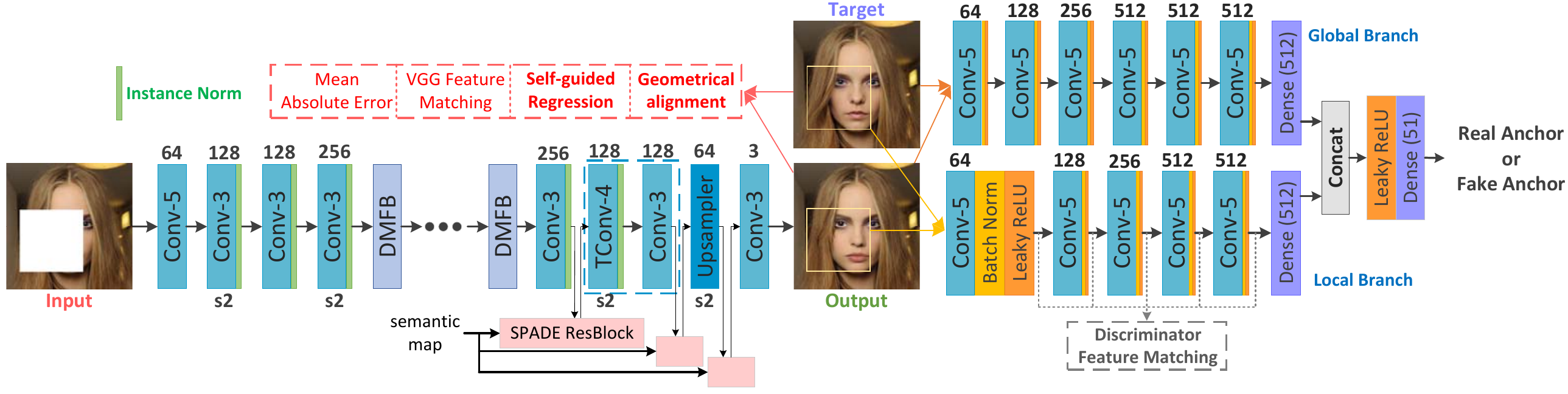}
            \subcaption{}        
        \end{minipage}    
        \hfill\vline\hfill
        \begin{minipage}[c]{0.2\textwidth}
            \includegraphics[width=\textwidth]{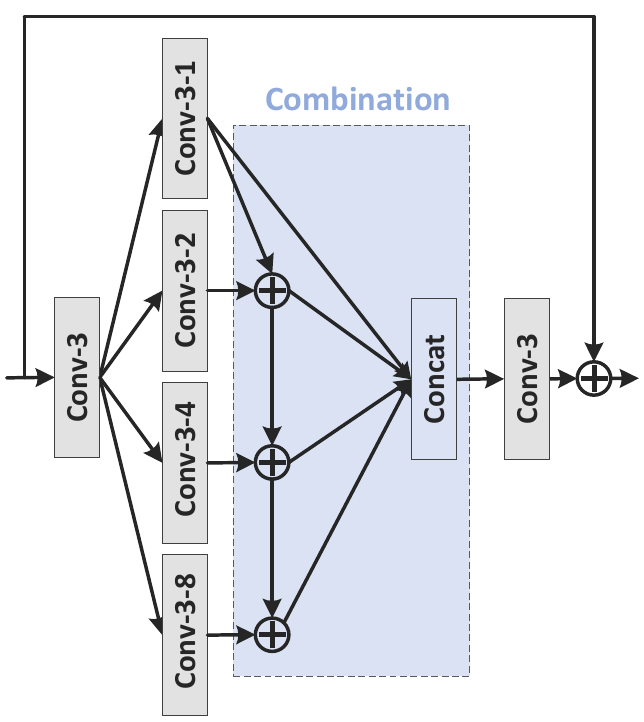}
            \subcaption{}
        \end{minipage}        
    \end{minipage}    
    \caption{Overview of Rainbow method. (a) The framework of the solution for classic image inpainting. (b) The framework of the solution for semantic guided image inpainting. (c) The structure of dense multi-scale fusion block (DMFB), ``Conv-3-8'' indicates $3 \times 3$ convolution layer with the dilation rate of 8.}
    \label{fig:rainbow}
\end{figure}

For both task 1 and task 2, Rainbow proposes a one-stage model (see Figure~\ref{fig:rainbow}) modified from Hui~\etal~\cite{hui2020image}, which utilizes dense combinations of dilated convolutions to obtain larger and more effective receptive fields. To better train the generator, and in addition to the standard VGG feature matching loss, they design a novel self-guided regression loss to dynamically correct low-level features of VGG guided by the current pixel-wise discrepancy map. Besides, they devise a geometrical alignment constraint item to penalize the coordinate center of estimated image high-level features away from the ground-truth. Moreover, for track 2, they add SPADE ResBlock~\cite{SPADE} in the decoder to introduce a semantic map for semantic guided image inpainting task.

With self-guided regression loss, geometrical alignment constraint, VGG feature matching loss, discriminator feature matching loss, realness adversarial loss~\cite{RealnessGAN}, and mean absolute error (MAE) loss, our overall loss function is defined as
\begin{equation}\label{eq:total_loss}
{{\cal L}_{total}} = {{\cal L}_{mae}} + \lambda \left( {{{\cal L}_{self - guided}} + {{\cal L}_{fm\_vgg}}} \right) + \eta {{\cal L}_{fm\_dis}} + \mu {{\cal L}_{real\_adv}} + \gamma {{\cal L}_{align}}.
\end{equation}
Except for ${{\cal L}_{real\_adv}}$, the detailed description of other losses can be found in~\cite{hui2020image}. Different from the standard GAN, RealnessGAN's discriminator outputs a distribution as the measure of realness.
\begin{equation}
{{\cal L}_{real\_adv}} = {\mathbb{E}_{z \sim {p_z}}}\left[ {{{\cal D}_{KL}}\left( {{{\cal A}_1}\left\| {D\left( {G\left( z \right)} \right)} \right.} \right)} \right] - {\mathbb{E}_{z \sim {p_z}}}\left[ {{{\cal D}_{KL}}\left( {{{\cal A}_0}\left\| {D\left( {G\left( z \right)} \right)} \right.} \right)} \right],
\end{equation}
where ${{{\cal D}_{KL}}\left( { \cdot \left\|  \cdot  \right.} \right)}$ denotes Kullback-Leibler (KL) divergence. Two virtual ground-truth distributions are needed to stand for the realness distributions of real and fakes images. They refer to these two distributions as ${{{\cal A}_1}}$ (real) and ${{{\cal A}_0}}$ (fake). Concretely, ${{{\cal A}_1}}$ and ${{{\cal A}_0}}$ are chosen to resemble the shapes of a Gaussian distribution $\cal{N}\left( {\mathbf{0},\mathbf{I}} \right)$ and a Uniform distribution $\cal{U}\left( {\mathbf{0},\mathbf{I}} \right)$, respectively.

It is set $\lambda  = 10$, $\eta  = 5$, and $\gamma  = 1$ in Equation~\ref{eq:total_loss}. The training procedure is optimized using Adam optimizer with ${\beta _1} = 0.5$ and ${\beta _2} = 0.9$. Learning rate $2e - 4$ , and batch size is $16$. The training patches are $256 \times 256$. For training, given a raw image ${\mathbf{I}_{gt}}$, a binary image mask $\mathbf{M}$ (value $0$ for known pixels and $1$ denotes unknown ones). In this way, the input image ${\mathbf{I}_{in}}$ is obtained from the raw image as ${\mathbf{I}_{in}} = {\mathbf{I}_{gt}} \odot \left( {1 - \mathbf{M}} \right)$. Our inpainting generator takes $\left[ {{\mathbf{I}_{in}},\mathbf{M}} \right]$ as input, and produces prediction ${\mathbf{I}_{pred}}$. The final output image is ${\mathbf{I}_{out}} = {\mathbf{I}_{in}} + {\mathbf{I}_{pred}} \odot \mathbf{M}$. All input and output are linearly scaled to $\left[ { - 1,1} \right]$. For arbitrary masks, the random regular regions are cropped and set to the local discriminator.

During the inference stage, our network cannot directly process input ultra-high-resolution images (up to $3872 \times 2592$px) with regular masks. When convolution operation is performed on masked regions, it cannot utilize features on known areas due to the large masked patches and the limited receptive field of the proposed generator. Thus, they downscale the input images and then generate low-resolution painted results. After that, bicubic interpolation is employed to produce high-resolution images. For irregular masks, it can regard as a low-level vision task. Because of the complete image with ultra-high-resolution, they crop it into four overlapped patches and then send them to the generator sequentially for solving the issue of OOM (out of memory).

\subsection{Yonsei-MVPLab}

\begin{figure}[t]
    \centering%
    \includegraphics[width=\textwidth, height=5.5cm]{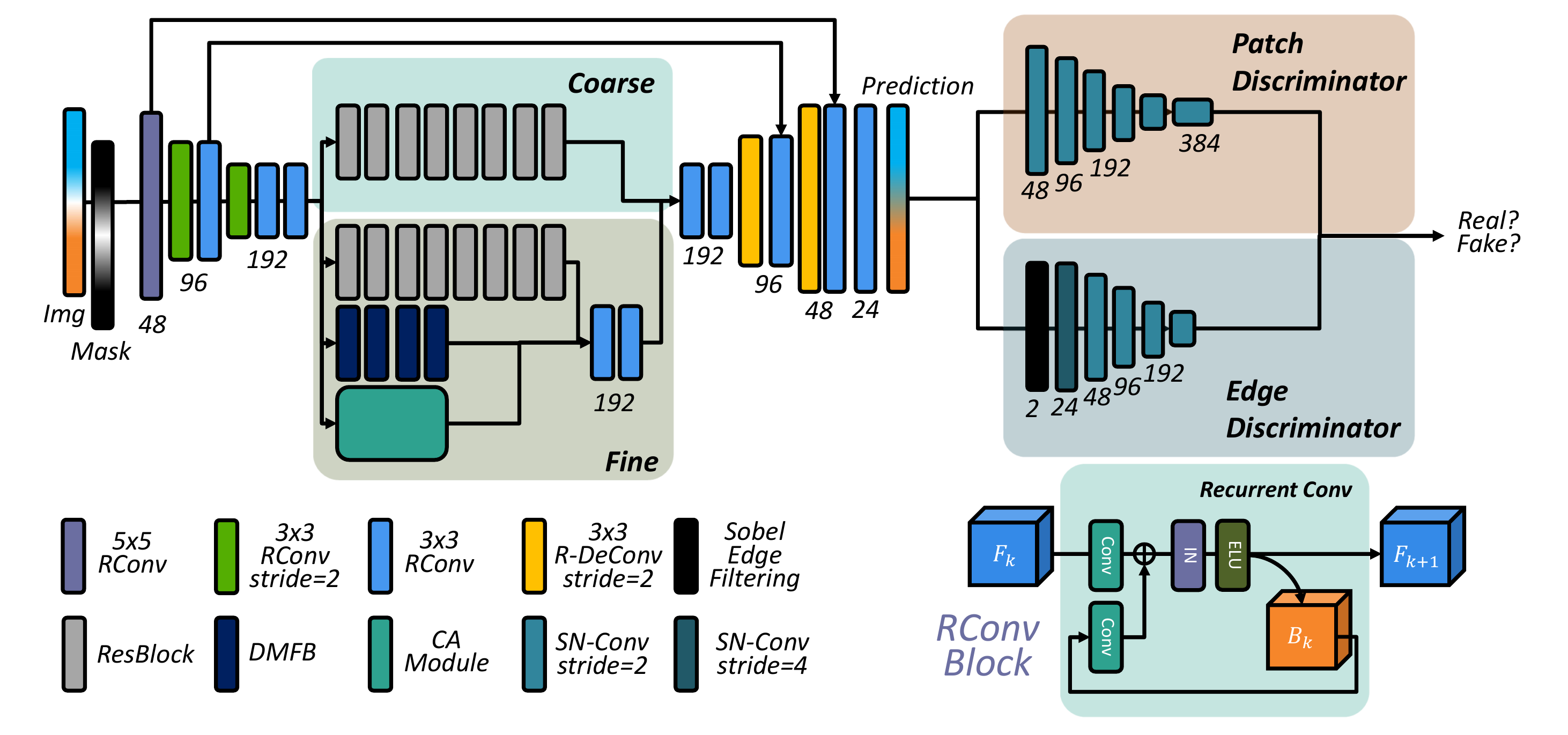}
    \caption{Overall network structures of Yonsei-MVPLab method. Basic building block is Recurrent Convolution (RConv) that is depicted in the lower-right side.}
    \label{fig:Yonsei-MVPLab}
\end{figure}

This team proposes three novel components for the task of image inpainting: Recurrent Convolution (RConv), Edge Discriminator, and Frequency Guidance Loss.
The overall structure is shown in~\fref{fig:Yonsei-MVPLab}.
RConv and R-DeConv indicate Recurrent Convolution and Recurrent Deconvolution, respectively.
DMFB and CA Module indicate Multi-scale Fusion Block in~\cite{hui2020image}, and contextual attention module in~\cite{yu2018generative}, respectively.
They follow the coarse-to-fine two-stage scheme as in similar inpainting algorithms\cite{yu2018generative,yu2019free}.
For the coarse route, residual blocks with dilated convolutions used in EdgeConnect~\cite{nazeri2019edgeconnect} are stacked.
For the fine route, the same residual blocks and additional parallel blocks: DMFB~\cite{hui2020image} and CA~\cite{yu2018generative} modules are used.

Like GatedConv~\cite{yu2019free}, Recurrent Convolution is a basic building block for every layer.
The recurrent Convolution block is depicted in the lower-right side of Fig.~\ref{fig:Yonsei-MVPLab}.
It saves the feature of a coarse route into memory and fusion with the feature from the fine route.
It effectively delivers gradients from the last output side of the network to the very first coarse layers via memory.  
For the discriminator, they modify the patch-based discriminator used in GatedConv~\cite{yu2019free}.
Therefore, a novel discriminator, called Edge Discriminator, is applied after the Sobel edge filter.
Sobel edge filter is implemented as convolution without gradient update that extracts horizontal and vertical edges.
These edges are passed into the edge discriminator.
It is much simple and not depend on the results of the predicted edges as in~\cite{nazeri2019edgeconnect}.
For our novel Frequency Guidance Losses, which is inspired by~\cite{fritsche2019frequency}, They filter the coarse output using a low-pass filter and the fine output using a high-pass filter. Then compare them with the same filtered ground truth images. For the coarse route, an L1 loss is applied to the low-pass filtered images, and for the fine route, an L1 loss is applied to the high-pass filtered images only in the hole region.

For some large holes, a selective up-scaling network is used for these images. These images are selected via convolution onto masks. For the up-scaling network, RRDBs in ESRGAN~\cite{wang2018esrgan} with 48 channels are used after the main network.
The hinge loss~\cite{lim2017geometric} is employed as the adversarial loss to train the discriminators.
For the generator, a total of eight losses are used. For the coarse rouse: l1 loss, adversarial loss, and perceptual loss~\cite{johnson2016perceptual}. For the fine route: style loss, feature matching loss, and two frequency guidance losses. Style loss and feature matching loss are borrowed from those of EdgeConnect~\cite{nazeri2019edgeconnect}.
The system is pretrained on Places dataset~\cite{zhou2017places}, and then fine-tune on the ADE20K~\cite{zhou2017scene} dataset. Training patch size is 256 $\times$ 256 (480 $\times$ 480 for up-scaling network). The learning rate for discriminator is 0.0004, and the learning rate for the generator and up-scaling network is 0.0001. The learning rate is decayed every 25 epochs by half. The entire system is trained for 100 epochs.

\subsection{BossGao}

\begin{figure}[t]
\centering%
\includegraphics[width=\textwidth]{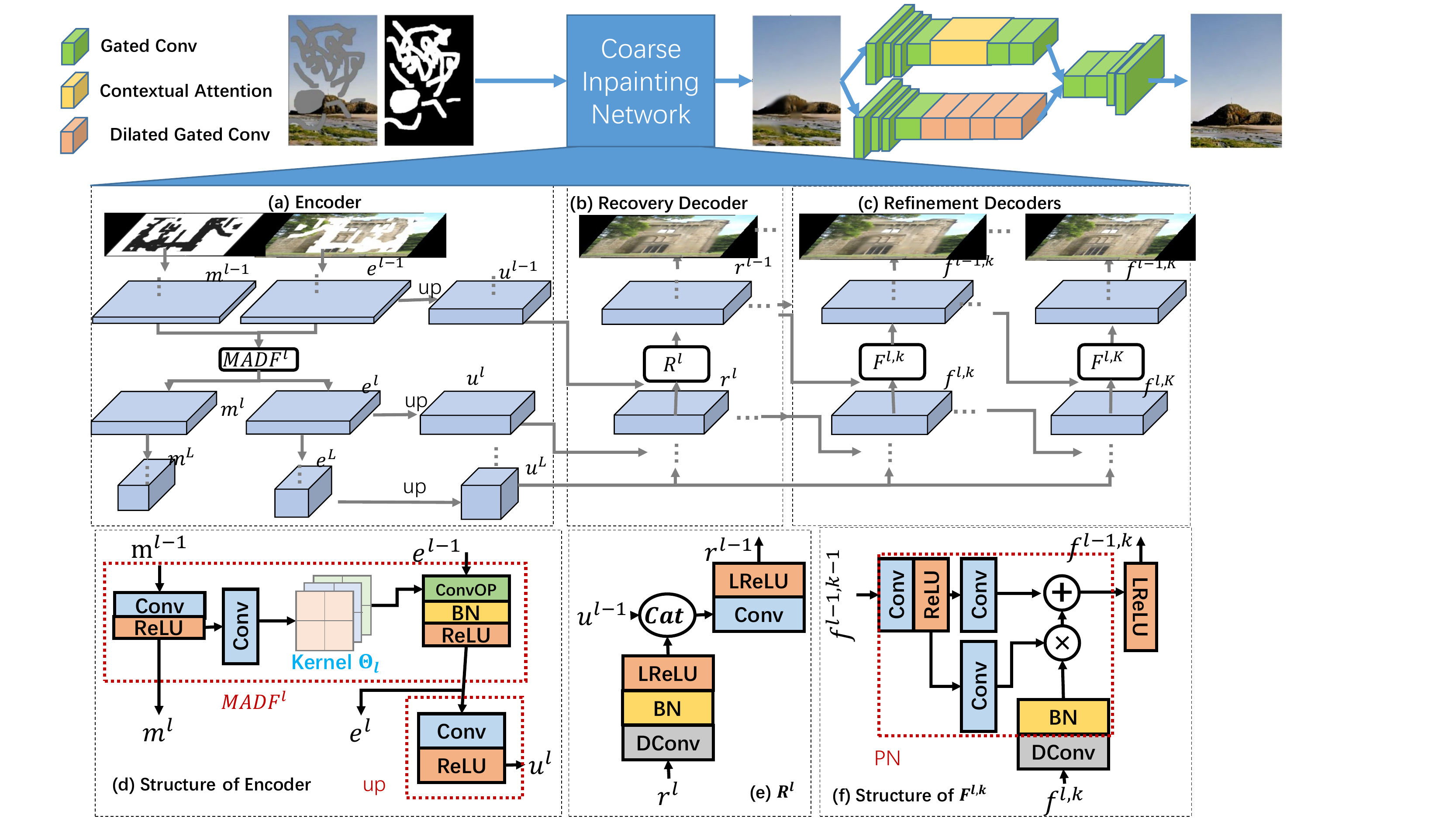}%
\caption{Two stage framework overview, Image Inpainting With Mask Awareness. Stage 1 fills missing regions coarsely and stage 2 refines the results. In this implementation, they use the exact stage 2 network configuration as DeepFill\_v2~\cite{yu2019free}. 
The architecture of the stage 1 network used for block mask is shown in the bottom. $MADF^l$ and $R^l$ are the MADF module and recovery decoder block at $l$-th level respectively. $F^{l,k}$ represents the $k$-th refinement decoder block at $l$-th level. $m^l$ is the $l$-th mask feature map of the encoder. The convolution operation marked in green in the encoder takes $e^{l-1}$ as input and its kernel for each convolution window is generated from the corresponding region of $m^{l-1}$.}
\label{fig:BossGao}
\end{figure}

In general, BossGao has designed two-stage solutions for extreme image inpainting, as shown in~\fref{fig:BossGao}. The first stage tries to recover coarse inpainting results which are relatively smooth for missing regions while keeping the content as semantically similar as possible. Our stage 2 uses the extract framework of DeepFill\_v2~\cite{yu2019free}, where gated convolution is designed to assigning soft attention scales to each feature point for adaptively modulation of different regions whose different shapes of valid points. In addition, contextual attention is adopted to leverage valid regions for better hallucination. In this challenge, our proposed image inpainting solution exhibits mask awareness in the following aspects: 1) they develop different networks to handle block mask, brush mask and cellural automata mask, respectively. 2) To handle extreme large block masks, they proposed a novel mask aware dynamic filtering (MADF) coarse inpainting network as stage 1. In MADF, filters for each convolution window are generated from features of the corresponding region of the mask. Furthermore, it is a 7-level downsample-upsample UNet architecture and is capable to generate relatively better coarse results than the coarse inpainting network of DeepFill\_{v2} because of its larger receptive field. With this solution, block masks can be filled in with much more visually plausible results, especially when the image is of very high resolution and the block mask is relatively large.

The details of stage 1 coarse network for box masks are as follows. The encoder $E$ generates multiple level feature maps. Let denote the finest level feature map as $u^1$ and the coarsest feature map as $u^L$, and $m^l$ and $e^l$ are the corresponding feature maps of mask and input generated by the encoder at level $l$.  
Let assume a $k\times k$ convolution with stride $s$ will be applied to $e^{l-1}\in R^{H\times W \times C_e^{l-1}}$, and there are $N_H\times N_W$ convolution windows in total. The mask-aware dynamic convolution layer marked in green operates as follows: on $m^{l-1}\in R^{H\times W \times C_m^{l-1}}$, $k\times k$ convolution with stride $s$ and ReLU activation is firstly applied to generate $m^l\in R^{N_H\times N_W\times C_m^l}$, then they utilize $1\times1$ convolution to generate the kernel tensor $\Theta_l\in R^{N_H\times N_W\times D}$, where $D$ equals $C_e^{l-1}\times k \times k\times C_e^l$. Finally, each of all the $N_H\times N_W$ windows in $e^{l-1}$ is convolved using the kernel reshaped from the corresponding point of $\Theta_l$, \textit{i.e.}, the convolution kernel of the $[n_H,n_W]$-th window in $e^{l-1}$ is reshaped from $\Theta_l[n_H,n_W,:]$.
To reduce computational overhead, they choose to apply MADF to extract relatively low dimensional latent features and then map it to a higher dimension. Specifically, $C_m^l$ is set to a relatively small value of 16 and output channel number $C_e^l$ is limited by $\min(128, 16*2^l)$ for any $l$. Then they increase the channels of each $e^l$ by $1\times1$ convolution with ReLU to produce $u^l$. Multiple cascaded decoders are used to refine the output step by step. 

\subsection{ArtIst}
\begin{figure}[t]
    \centering%
    \includegraphics[width=0.9\textwidth]{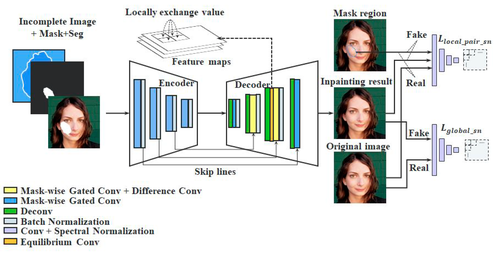}
    \caption{Overall architecture of the ArtIst approach.}
    \label{fig:artist}
\end{figure}

This solution proposes a one-stage inpainting solution~\cite{bai2020deepGIN} based on a single UNet architecture consisting of 6 different kinds of layers (operations), where the main contribution is different convolution-based blocks. First, Equilibrium Convolution (EConv) is designed to reduce undesired artifacts in the output image. Second, Mask-wise Convolutions (MGConv) are deployed to reduce the parameters in gated convolutions~\cite{yu2019free}. Lastly, in order to deal with the large mask challenge, they create a Differentiate Convolution block. A schematic of the architecture is depicted in \fref{fig:artist}.

\subsection{DLUT}
\begin{figure}[t]
\centering%
\includegraphics[width=.6\textwidth, height=4cm]{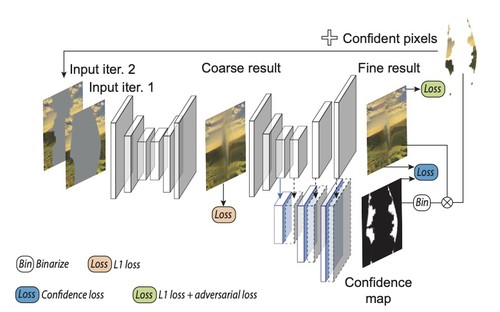}\includegraphics[width=.4\textwidth, height=4cm]{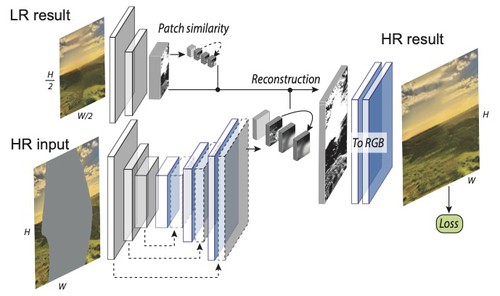}
\caption{Overview of DLUT method. }
\label{fig:dlut}
\end{figure}

This method is mainly based on the inpainting system proposed by Zeng~\etal~\cite{zeng2020high}. It is an iterative inpainting model with a guided upsampling module as shown in~\fref{fig:dlut}.
Due to the memory limitation, for high-resolution images whose the long side is greater than a threshold, they execute the inpainting system of Zeng~\etal\cite{zeng2020high} on $2\times$ downsampled input and then use an external super-resolution model~\cite{tfsr}\footnote{https://github.com/tensorlayer/srgan.git} to get the inpainted images of the original size.

\subsection{AI-Inpainting}

\begin{figure}[t]
    \centering%
    \includegraphics[width=\textwidth, height=5.5cm]{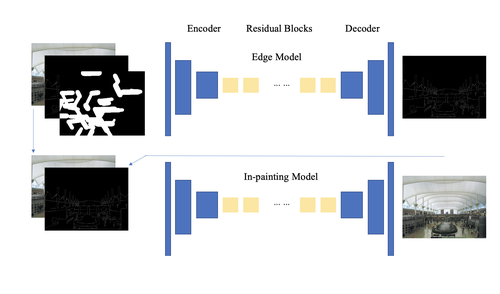}
    \caption{Overview of AiriaBeijing method. The first module predicts semantic edges, the second uses these edges to estimate the results (zoom-in for better details).}
    \label{fig:airiabeijing}
\end{figure}

The solution is based on Edge-Connect~\cite{Nazeri_2019_ICCV}, which is mainly focused on small size images. The entire network consists of two sub-modules: an edge part and an in-painting part (see Figure~\ref{fig:airiabeijing}). The edge module uses the masked image to generate the edge of the entire image, and the edge together with the masked image are fed into the inpainting module in order to generate the final result. This system categorizes images into three different groups according to their mask type. For box masks, the image will be resized to three smaller sizes in order to be filled up with details at different aspects, and thus, be able to run the model without running out of memory. After resizing back to original size, they merge three images with different weight into one image. For large images with other masks, they cut each image into many patches and fed them into the model. After all, they put the output images back to original position. This method might cause some color deviation between masks since they do not use global information of the whole image. 

\subsection{qwq}
\begin{figure}[t]
\centering%
\includegraphics[width=\textwidth, height=4cm]{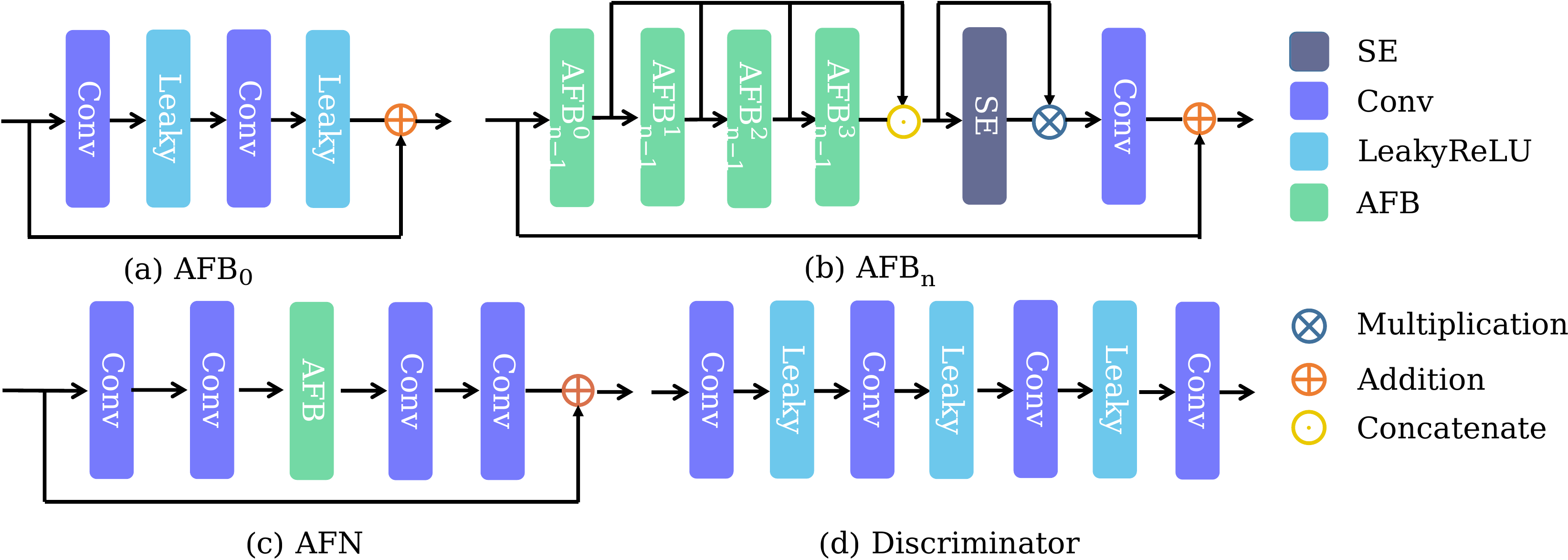}%
\caption{Architecture of the qwq proposed framework. (a) shows the structure of the basic block $\text{AFB}_0$. (b) shows the hierarchical
architecture of $\text{AFB}_n$, where $n\in [1,r]$ is the recursion level of the block. (c) illustrates the structure of AFN, which consists
of encoding layers, an $\text{AFB}_r$, and decoding layers from left to right. (d) shows the structure of our discriminator.}
\label{fig:qwq}
\end{figure}

The proposed solution adopts a AFN network~\cite{Xu_2020_CVPR_Workshops} as generator, and a Markovian Discriminator~\cite{li2016precomputed} to guide the inpainting process. As shown in~\fref{fig:qwq}, the generator consists of several fractally stacked Attentive Fractal
Blocks, which are constructed via progressive feature fusion and channel-wise attention guidance. Shortcuts and residual
connections at different scales effectively resolve the vanishing gradients and help the network to learn more key features.
The progressive fusion of intermediate features lets the network handle rich information. The discriminator is constructed via stacking several convolution and leaky ReLU layers. The output of the discriminator is designed to be a one-channel confidence map, which penalizes the masked area.

\subsection{CVIP Inpainting Team}
\begin{figure}[t]
    \centering%
    \includegraphics[width=0.9\textwidth]{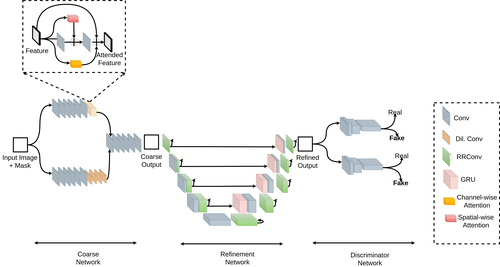}
    \caption{Overall architecture of the CVIP approach. The model has two stages, namely a coarse network and a refinement network.}
    \label{fig:cvip}
\end{figure}

Following the work of Yu \textit{et al.}~\cite{yu2018generative}, this team proposes a model consisting of a two-stage system, namely a coarse network, and a refinement network. \fref{fig:cvip} shows the proposed solution.
In addition to a regular branch, the coarse network uses an attention branch with a novel attention module namely the Global Spatial-Channel Attention (GSCA) module, which can capture the structural consistency by calculating global correlation among both spatial and channel features at the global level. For the refinement network, in addition to GSCA, they adopt a recurrent residual attention U-net (RR2U)~\cite{alom2018recurrent} architecture with major modifications in the architectural level. The recurrent residual attention consists of a multi-stage GRUs~\cite{cho2014learning} that computes the inter-layer feature dependencies.

\subsection{DeepInpaintingT1}
\begin{figure}[t]
\centering%
\includegraphics[width=0.95\textwidth, height=4cm]{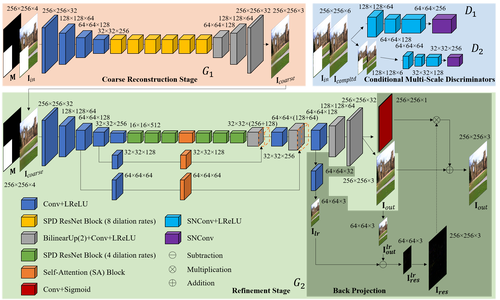}%
\caption{Overview of DeepInpaintingT1 for extreme image inpainting. }
\label{fig:DeepInpaintingT1}
\end{figure}

The proposed model~\cite{li2020deepGIN} consists of two generators and two discriminators. The overall proposed architecture is displayed in \fref{fig:DeepInpaintingT1}. The first coarse generator $\textit{G}_{1}$ at Coarse Reconstruction Stage (the top left corner) and the second refinement generator $\textit{G}_{2}$ at the Refinement Stage (bottom) constitute the Deep Generative Inpainting Network (DeepGIN) which is used in both training and testing. The two discriminators $\textit{D}_{1}$ and $\textit{D}_{2}$ located within Conditional Multi-Scale Discriminators area (the top right corner) are only used in training as an auxiliary network for generative adversarial training. The generator $\textit{G}_{1}$ is trained to roughly reconstruct the missing regions and gives $\textbf{I}_{coarse}$. The generator $\textit{G}_{2}$ is trained to exquisitely decorate the coarse prediction with details and textures, and eventually forms the completed image $\textbf{I}_{out}$. Three main ideas are proposed to accomplish the task of inpainting: \textit{i}) Spatial Pyramid Dilation ResNet blocks (yellow and green blocks) with various dilation rates to enlarge the receptive fields such that information given by distant spatial locations can be used for both reconstruction and refinement; \textit{ii}) Multi-Scale Self Attention (orange blocks) to enhance the coherency of the completed image by attending on the self-similarity of the image itself at three different scales; \textit{iii}) Back Projection strategy (the bottom right shaded area) to encourage better alignment of the generated patterns and the reference ground truth. For the two discriminators (the top right corner), they operate at two input scales, 256$\times$256 and 128$\times$128 respectively, to encourage better details and textures of the locally generated patterns at different scales.

\subsection{IPCV\_IITM}

\begin{figure}[t]
    \centering%
    \includegraphics[width=\textwidth, height=5.5cm]{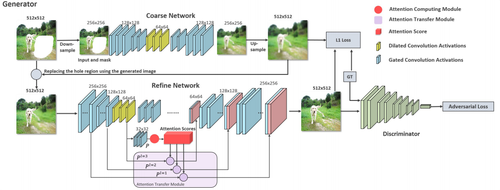}
    \caption{Overview of IPCV\_IITM method. Two models run in parallel to generate results that is evaluated using a discriminator module.}
    \label{fig:ipcv_iitm}
\end{figure}

Team IPCV\_IITM proposes a method based on Contextual Residual Aggregation (CRA) network~\cite{yi2020contextual}. Although recent deep learning-based in-painting methods are more effective than classic approaches, however, due to memory limitations they can only handle low-resolution inputs, typically smaller than 1K. Meanwhile, the resolution of photos captured with mobile devices increases up to 8K. Naive up-sampling of the low-resolution inpainted result can merely yield a large yet blurry result. Whereas, adding a high-frequency residual image onto the large blurry image can generate a sharp result, rich in details and textures. Motivated by this, CRA mechanism is employed that can produce high-frequency residuals for missing contents by weighted aggregating residuals from contextual patches, thus only requiring a low-resolution prediction from the network.
\fref{fig:ipcv_iitm} shows the network design proposed by this team.
Since convolutional layers of the neural network only
need to operate on low-resolution inputs and outputs, the cost of memory and computing power is thus well suppressed.

\subsection{MultiCog}
\begin{figure}[t]
\centering%
\includegraphics[width=\textwidth, height=4cm]{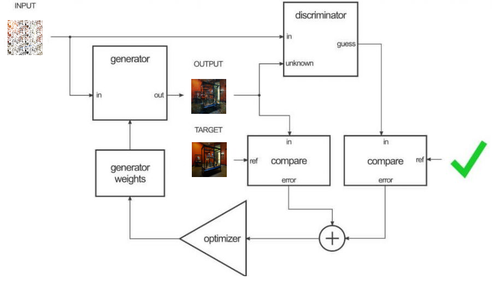}%
\caption{Overview of MultiCog.}
\label{fig:MultiCog}
\end{figure}

MultiCog uses a Pix2Pix Generative Adversarial Network (GAN) as its core~\cite{isola2016imagetoimage}. The main idea here is selecting the two domains as unpainted images and painted images, where the learned mapping function is unpainted to painted. See schematic in \fref{fig:MultiCog}.




\section*{Acknowledgements}
We thank the AIM 2020 sponsors: Huawei, MediaTek, Qualcomm AI Research, NVIDIA, Google and Computer Vision Lab / ETH Z\"urich.
\section*{Appendix A: Teams and affiliations}
\label{sec:affiliation}

\subsection*{AIM2020 organizers}
\textbf{Members}: Evangelos Ntavelis\textsuperscript{1,2} (entavelis@ethz.ch), Siavash Bigdeli\textsuperscript{2} (siavash.bigdeli@csem.ch), Andr\'es Romero\textsuperscript{1} (roandres@ethz.ch), Radu Timofte\textsuperscript{1} (radu.timofte@vision.ee.ethz.ch).
\noindent\textbf{Affiliations}: \textsuperscript{1}Computer Vision Lab, ETH Z\"urich. \textsuperscript{2}CSEM.

\subsection*{Rainbow}
\textbf{Title}: Image fine-grained inpainting.

\noindent\textbf{Members}: Zheng Hui, Xiumei Wang, Xinbo Gao.

\noindent\textbf{Affiliations}: School of Electronic Engineering, Xidian University.

\subsection*{Yonsei-MVPLab}
\textbf{Title}: Image Inpainting based on Edge and Frequency Guided Recurrent Convolutions.

\noindent\textbf{Members}: Chajin Shin, Taeoh Kim, Hanbin Son, Sangyoun Lee.

\noindent\textbf{Affiliations}: Image and Video Pattern Recognition Lab., School of Electrical and Electronic Engineering, Yonsei University, Seoul, South Korea.

\subsection*{BossGao}
\textbf{Title}: Image Inpainting With Mask Awareness

\noindent\textbf{Members}: Chao Li, Fu Li, Dongliang He, Shilei Wen, Errui Ding

\noindent\textbf{Affiliations}: Department of Computer Vision (VIS), Baidu Inc.

\subsection*{ArtIst}
\textbf{Title}: Fast Light-Weight Network for Image Inpainting

\noindent\textbf{Members}: Mengmeng Bai, Shuchen Li

\noindent\textbf{Affiliations}: Samsung R\&D Institute China-Beijing (SRC-Beijing)

\subsection*{DLUT}
\noindent\textbf{Title}: Iterative Confidence Feedback and Guided Upsampling for filling large holes and inpainting high-resolution images

\noindent\textbf{Members}: Yu Zeng\textsuperscript{1}, Zhe Lin\textsuperscript{2}, Jimei Yang\textsuperscript{2}, Jianming Zhang\textsuperscript{2}, Eli Shechtman\textsuperscript{2}, Huchuan Lu\textsuperscript{1}

\noindent\textbf{Affiliations}: \textsuperscript{1}Dalian University of Technology, \textsuperscript{2}Adobe

\subsection*{AI-Inpainting Group}
\textbf{Title}: MSEM: Multi-Scale Semantic-Edge Merged Model for Image Inpainting

\noindent\textbf{Members}: Weijian Zeng, Haopeng Ni, Yiyang Cai, Chenghua Li

\noindent\textbf{Affiliations}: Rensselaer Polytechnic Institute

\subsection*{qwq}
\textbf{Title}: Markovian Discriminator guided Attentive Fractal Network

\noindent\textbf{Members}: Dejia Xu, Haoning Wu, Yu Han

\noindent\textbf{Affiliations}: Peking University

\subsection*{CVIP Inpainting Team}
\textbf{Title}: Global Spatial-Channel Attention and Inter-layer GRU-based Image Inpainting

\noindent\textbf{Members}: Uddin S. M. Nadim, Hae Woong Jang, Soikat Hasan Ahmed, Jungmin Yoon, and Yong Ju Jung

\noindent\textbf{Affiliations}: Computer Vision and Image Processing (CVIP) Lab, Gachon University.

\subsection*{DeepInpaintingT1}
\textbf{Title}: Deep Generative Inpainting Network for Extreme Image Inpainting

\noindent\textbf{Members}: Chu-Tak Li, Zhi-Song Liu, Li-Wen Wang, Wan-Chi Siu, Daniel P.K. Lun

\noindent\textbf{Affiliations}: Centre for Multimedia Signal Processing, Department of Electronic and
Information Engineering, The Hong Kong Polytechnic University, Hong
Kong

\subsection*{IPCV IITM}
\textbf{Title}: Contextual Residual Aggregation Network

\noindent\textbf{Members}: Maitreya Suin, Kuldeep Purohit, A. N. Rajagopalan

\noindent\textbf{Affiliations}: Indian Institute of Technology Madras, India

\subsection*{MultiCog}
\textbf{Title}: Pix2Pix for Image Inpainting

\noindent\textbf{Members}: Pratik Narang\textsuperscript{1}, Murari Mandal\textsuperscript{2}, Pranjal Singh Chauhan\textsuperscript{1}

\noindent\textbf{Affiliations}: \textsuperscript{1}BITS Pilani, \textsuperscript{2}MNIT Jaipur


%
%
\bibliographystyle{splncs04}
\bibliography{egbib}
\end{document}